\documentclass[a4paper,fleqn]{cas-dc}

\usepackage[authoryear]{natbib}
\usepackage{float}
\usepackage{graphicx}
\usepackage{placeins}
\usepackage{xcolor}
\usepackage{url}
\usepackage{algorithm}
\usepackage{algpseudocode}
\usepackage{placeins}
\usepackage{subcaption}
\usepackage{tikz}
\usepackage{pgfplots}
\pgfplotsset{compat=1.18}
\usepackage{booktabs}
\usepackage{tabularx}
\usepackage{array}
\usepackage{multirow}
\usepackage{booktabs}
\usepackage{threeparttable}
\usepackage{makecell}

\def\tsc#1{\csdef{#1}{\textsc{\lowercase{#1}}\xspace}}
\tsc{WGM}
\tsc{QE}
\tsc{EP}
\tsc{PMS}
\tsc{BEC}
\tsc{DE}

\begin{document}

\let\WriteBookmarks\relax

\def\textpagefraction{.001}
\shorttitle{Boundary-Balanced Replay for Continual Segmentation}
\shortauthors{Zahid et~al.}

\title [mode = title]{BBR-Net: Boundary-Balanced Replay for Continual Medical Image Segmentation}                      



\author[1]{Zahid Ullah}[orcid=0000-0002-0184-7620]
\credit{Conceptualization, Methodology, Software, Formal analysis, Investigation, Data curation, Writing - original draft, Writing - review \& editing.}

\author[1]{Sieun Choi}[orcid=0009-0006-6623-4645]
\credit{Conceptualization, Methodology, Formal analysis, Writing - original draft, Writing - review \& editing.}

\author[1]{Jihie Kim}[orcid=0000-0003-2358-4021]
\cormark[1]
\ead{jihie.kim@dgu.edu}
\credit{Formal analysis, Investigation, Supervision, Project administration, Project management.}

\affiliation[1]{organization={Department of Computer Science and Artificial Intelligence},
                addressline={Dongguk University},
                city={Seoul},
                postcode={04620},
                country={Republic of Korea}}

\cortext[cor1]{Corresponding author}


\begin{abstract}
Continual learning for medical image segmentation remains challenging under domain shift because replay-based methods often preserve appearance information without explicitly modeling anatomical structure. This study investigates whether structural consistency governs knowledge retention in continual cardiac ultrasound segmentation. We propose the Boundary-Balanced Replay Network (BBR-Net), which selects replay samples using boundary-aware priority and class balance to preserve anatomically informative regions. The method is evaluated on CAMUS and CardiacNet under forward (CAMUS to CardiacNet) and reverse (CardiacNet to CAMUS) task orders. In the forward setting, BBR-Net retains source-task performance close to an offline joint-training reference, while markedly reducing catastrophic forgetting and preserving competitive target-task adaptation. Ablation results show that boundary-aware prioritization contributes to retention and improves the balance between source-task preservation and target-task adaptation when combined with class-aware sampling. In contrast, the reverse setting reveals that structure-aware replay fails when initial representations are learned from noisy and structurally inconsistent data. To isolate this effect, we conduct a controlled structural perturbation analysis by progressively corrupting source-task boundaries while keeping the dataset, architecture, and training protocol fixed. Forgetting increases consistently as structural reliability decreases, suggesting that replay effectiveness is strongly influenced by the quality of stored structural information, rather than by memory capacity alone. These findings indicate that preserving anatomical structure under domain shift is a central factor in continual medical image segmentation, and that replay mechanisms should account for structural reliability to support robust knowledge retention. The code is available at: \url{https://github.com/Zahid672/Continual_Learning_BBR-Net}
\end{abstract}




\maketitle

\section{Introduction}
\label{intro}

Deep learning has transformed medical image analysis, particularly in cardiac imaging \cite{mazher2024self,yu2023multi}, where accurate delineation of anatomical structures is essential for diagnosis, intervention planning, and disease monitoring \cite{comaniciu2016shaping,muffoletto2025neural}. Over the past decade, encoder–decoder architectures, most notably U-Net and its variants, have achieved remarkable performance in segmenting cardiac structures from ultrasound and MRI data \cite{ronneberger2015unet, isensee2021nnu}. These models, however, are typically developed under a simplifying assumption: that all training data are available at once, drawn from a consistent distribution, and curated under uniform acquisition protocols. Clinical practice does not conform to this assumption. Data are inherently fragmented, acquired over time across institutions, devices, and patient populations, and often restricted by privacy and regulatory constraints \cite{parisi2019continual,delange2021continual}. As a result, models must be updated sequentially rather than retrained from scratch, raising the need for continual learning (CL) in medical imaging \cite{wu2024continual,qazi2024medicalclsurvey}.

CL seeks to enable models to adapt to new data while retaining previously acquired knowledge, thereby avoiding catastrophic forgetting \cite{parisi2019continual, delange2021continual}. In principle, an ideal continual segmentation model would accumulate knowledge over time, preserving anatomically meaningful representations regardless of shifts in image appearance \cite{kumari2025continual,zhu2024boosting}. It would recognize cardiac structures consistently, whether the input originates from a clean benchmark dataset or a noisy clinical environment. In practice, however, existing CL methods fall short of this ideal. When trained sequentially, segmentation models tend to forget previously learned structures, particularly when the data distribution changes significantly between tasks \cite{zhang2022rcil,wu2024continual}. This challenge becomes more severe in medical image segmentation, where domain shifts caused by differences in imaging devices, acquisition protocols, and annotation quality can substantially alter appearance while preserving underlying anatomical structure \cite{qazi2024medicalclsurvey,perkonigg2021dynamic}.

A large body of CL research has attempted to address this issue through regularization, architectural expansion, and replay-based strategies. Regularization-based methods constrain parameter updates to protect earlier knowledge \cite{kirkpatrick2017ewc}, while dynamic architectures allocate task-specific capacity to reduce interference \cite{rusu2016progressive}. Among these, replay-based approaches remain the most widely adopted, especially in vision tasks, as they reintroduce samples from prior tasks during training \cite{rebuffi2017icarl}. Yet, despite their practical success, these methods implicitly assume that preserving appearance-level information, pixel distributions or feature embeddings is sufficient to retain knowledge. This assumption may hold for natural image classification, where appearance is often discriminative, but it becomes problematic in medical image segmentation.

\begin{figure}
    \centering
    \includegraphics[width=0.9\columnwidth]{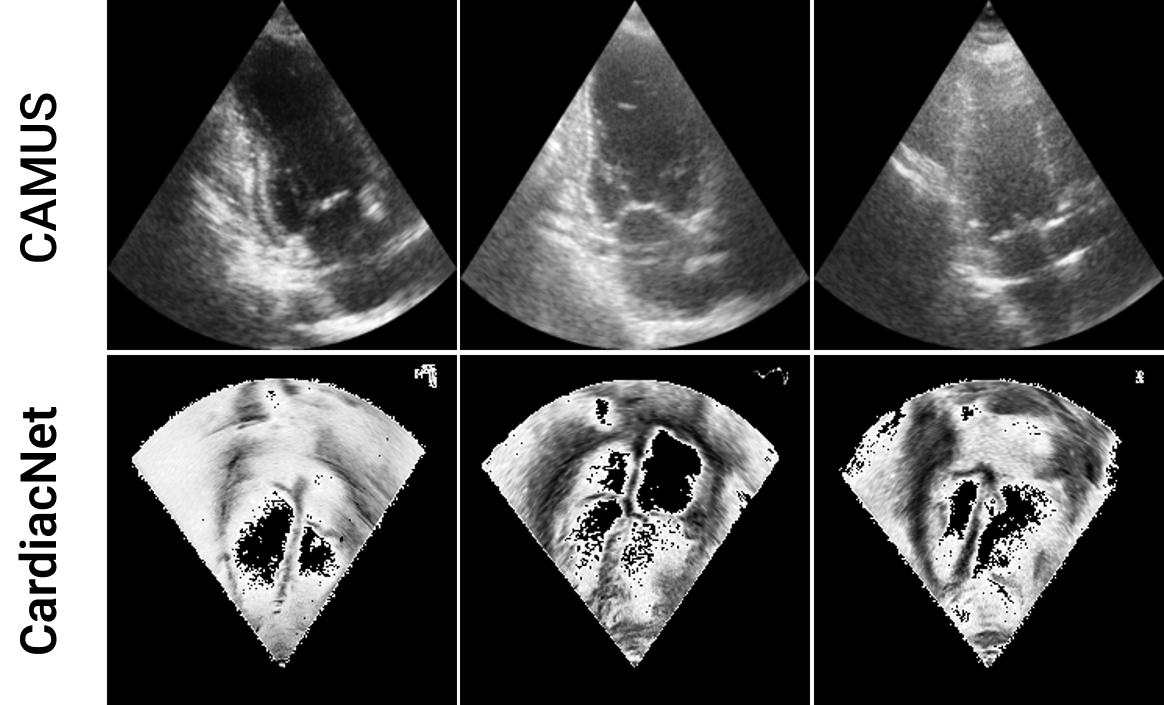}
    \caption{Sample echocardiographic images from the CAMUS (top row) and CardiacNet (bottom row) datasets. The CAMUS dataset contains relatively clean echocardiographic images with well-defined cardiac structures, whereas the CardiacNet dataset exhibits higher variability, including severe speckle noise, low contrast, and irregular anatomical boundaries.}
    \label{datasets_sample}
\end{figure}

As shown in Fig. \ref{datasets_sample}, cardiac imaging presents a different challenge. While image appearance varies substantially due to acquisition conditions, noise, and patient variability, the underlying anatomical structure remains relatively stable \cite{guan2022domain,perkonigg2021dynamic}. The geometry of the left ventricle, the boundaries of the myocardium, and the spatial relationships between chambers provide important structural cues for cardiac segmentation \cite{heimann2009statistical,kervadec2021boundary}. This suggests a fundamental mismatch in current CL approaches: they tend to prioritize appearance, which varies across domains, rather than anatomical structure, which remains relatively stable. As a consequence, models may appear to retain knowledge numerically while failing to preserve anatomically consistent representations. This disconnect is subtle but critical, as segmentation quality in clinical applications depends not only on region overlap but also on boundary accuracy and structural coherence \cite{litjens2017survey,kervadec2021boundary}.

This observation leads to the central problem addressed in this work. Existing replay-based CL methods do not explicitly account for anatomical structure, and therefore struggle under domain shift where appearance varies but structure remains invariant \cite{wu2024continual, rebuffi2017icarl}. The ideal scenario would involve a model that selectively retains structurally informative samples that encode boundaries and spatial relationships, while adapting to new domains. In reality, current methods treat all samples equally or prioritize them based on appearance-driven criteria, leading to suboptimal knowledge retention.

Several recent studies have begun to explore more sophisticated memory and sampling strategies for medical CL, including atypical replay sample selection, domain continual segmentation, and class-imbalance-aware memory construction \cite{bera_memory_2023,zhu2023memory,li2025clms,zhao2025rethinking}. While these approaches improve performance to some extent, they still operate within an appearance-centric framework and do not directly model structural properties of segmentation tasks. As a result, they fail to address a key question: what type of information should be preserved to ensure reliable knowledge retention in medical image segmentation?

The consequences of overlooking structure are both immediate and far-reaching. Directly, models exhibit degraded segmentation performance when transitioning across datasets, with fragmented predictions and inconsistent boundaries \cite{isensee2021nnu}.  Indirectly, this undermines the deployment of adaptive learning systems in clinical settings, where robustness across domains is essential \cite{litjens2017survey, wu2024continual}. A model that cannot maintain structural consistency over time cannot be trusted in longitudinal or multi-center applications, regardless of its performance on isolated benchmarks.

In this work, we take a structural perspective on CL. We hypothesize that anatomical consistency, rather than appearance similarity, is the key factor affecting knowledge retention in medical image segmentation. To test this hypothesis, we introduce a Boundary-Balanced Replay Network (BBR-Net), which incorporates structural information into memory selection by prioritizing boundary regions and enforcing class balance. Unlike conventional replay methods \cite{rebuffi2017icarl, chaudhry2019tiny}, BBR-Net explicitly emphasizes anatomically informative samples, aligning the learning process with the geometric nature of the task.

Beyond proposing a method, we conduct a systematic investigation of task-order effects under domain shift. Specifically, we analyze forward (CAMUS to CardiacNet) and reverse (CardiacNet to CAMUS) learning scenarios, revealing a striking asymmetry in performance. While structure-aware replay is highly effective when stable anatomical priors are learned early, it fails when initial representations are noisy and structurally inconsistent. This finding highlights a fundamental limitation of replay-based CL \cite{mnih2015dqn}: its effectiveness depends not only on the presence of memory, but also on the structural reliability of the representations stored in memory.

To further isolate this effect, we conduct a controlled structural perturbation experiment in which the structural reliability of source-task masks is progressively degraded before replay buffer construction, while keeping the dataset order, model architecture, training schedule, and memory size fixed. This experiment allows us to examine whether replay degradation is associated with structural corruption while reducing dataset-level confounders such as acquisition variability, task difficulty, or annotation noise. The observed increase in forgetting under stronger structural perturbation provides controlled empirical support that replay effectiveness is affected by the quality of stored structural information.

The objectives of this study are therefore fourfold. First, to examine whether structural consistency provides a more reliable foundation for CL than appearance similarity. Second, to develop a replay mechanism that integrates boundary and class-aware prioritization. Third, to analyze how task order influences learning dynamics and knowledge retention under domain shift. Fourth, to directly validate the structural consistency hypothesis through controlled perturbation of source-task anatomical labels. Together, these objectives aim to move beyond incremental performance improvements and toward a deeper understanding of what drives effective CL in medical imaging.

The significance of this work lies in its dual contribution. From a theoretical perspective, it challenges prevailing assumptions about representation preservation in CL. From a practical perspective, it provides insights into designing robust segmentation models capable of adapting across heterogeneous clinical environments. Rather than treating replay as a universal solution, our findings suggest that its success is conditional, and that structural reliability is a critical factor in determining its effectiveness.

The main contributions of this work are summarized as follows:
\begin{itemize}
    
    \item We propose BBR-Net, a boundary-balanced replay framework that integrates class-aware and boundary-aware prioritization for continual medical image segmentation.
    
    
    \item We demonstrate that structural reliability is a key factor affecting replay effectiveness, beyond memory size or sample frequency alone.
    
    \item We provide controlled structural perturbation evidence showing that progressive boundary degradation leads to systematically increased forgetting.

    \item We analyze task-order effects under domain shift and reveal a key limitation of structure-aware replay when initial structural priors are unreliable.
    
    
\end{itemize}

The remainder of this paper is organized as follows. Section~\ref{relatework} reviews related work on CL and medical image segmentation. Section~\ref{propose} describes the proposed BBR-Net framework. Section~\ref{experimental} presents the experimental setup, including datasets, implementation details, and evaluation metrics. Section~\ref{results} reports the quantitative, qualitative, ablation, and controlled structural perturbation results. Section~\ref{sec:discussion} discusses the main findings and their implications. Finally, Section~\ref{conclusion} concludes the paper and outlines future research directions.

\section{Related Work}
\label{relatework}

CL \cite{wang2024comprehensive,peng2025rethinking} has emerged as a fundamental paradigm for enabling models to adapt to non-stationary data distributions without catastrophic forgetting. Its importance is particularly pronounced in medical image segmentation \cite{xu2025multi,li2025clms}, where data are inherently sequential, heterogeneous, and constrained by privacy considerations. Recent surveys highlight that, despite significant progress, CL in medical imaging remains an open challenge due to domain shift, annotation variability, and limited access to historical data~\cite{wu2024continual, qazi2024medicalclsurvey}. In such settings, models must evolve over time while maintaining previously acquired knowledge, a requirement that remains only partially addressed by current methodologies. The objective of this study is to investigate whether structural consistency provides a more reliable foundation for CL than appearance-based strategies, to develop a structure-aware replay mechanism, and to analyze the influence of task order under domain shift. Table~\ref{tab:related_work} summarizes representative CL approaches related to medical image segmentation and positions BBR-Net with respect to these methods.

\subsection{Continual Learning}

Existing CL approaches can be broadly categorized into regularization-based and replay-based methods. Regularization-based approaches aim to preserve prior knowledge by constraining parameter updates. Elastic Weight Consolidation (EWC)~\cite{kirkpatrick2017ewc} introduces a quadratic penalty based on the Fisher information matrix to prevent significant changes to important parameters. Similarly, Learning without Forgetting (LwF)~\cite{li2017learning} maintains consistency with previous predictions through knowledge distillation. While these methods are effective in classification tasks, they exhibit limited applicability in dense prediction problems such as segmentation, where preserving spatially localized and structurally coherent information is essential. As noted in recent surveys~\cite{wu2024continual}, such parameter-level constraints are insufficient to capture the complex spatial dependencies inherent in medical imaging tasks.

Replay-based methods, in contrast, address forgetting by explicitly storing and revisiting samples from previous tasks. Representative approaches such as iCaRL~\cite{rebuffi2017icarl} and Tiny Episodic Memory~\cite{chaudhry2019tiny} have demonstrated strong performance in incremental learning scenarios. These methods operate under the assumption that revisiting past samples can approximate joint training. However, their effectiveness relies heavily on how representative the stored samples are. In most cases, replay strategies are either random or guided by simple heuristics, implicitly assuming that preserving appearance-level information is sufficient for knowledge retention. Recent analyses~\cite{qazi2024medicalclsurvey} suggest that such assumptions may not hold in domains where appearance varies significantly but underlying semantics remain stable.

\begin{table*}[!t]
\centering
\caption{Comparison with representative CL approaches related to medical image segmentation.}
\label{tab:related_work}
\footnotesize
\renewcommand{\arraystretch}{1.12}
\setlength{\tabcolsep}{4pt}
\begin{tabular*}{\linewidth}{
@{\extracolsep{\fill}}
>{\raggedright\arraybackslash}p{0.20\linewidth}
>{\raggedright\arraybackslash}p{0.30\linewidth}
>{\raggedright\arraybackslash}p{0.42\linewidth}
@{}}
\hline
\textbf{Category} & \textbf{Representative Work} & \textbf{Relation to BBR-Net} \\
\hline

Regularization-based CL
& EWC~\cite{kirkpatrick2017ewc}, LwF~\cite{li2017learning}
& Preserves parameters or predictions but does not explicitly model anatomical structure or replay quality. \\

Replay-based CL
& iCaRL~\cite{rebuffi2017icarl}, Tiny Memory~\cite{chaudhry2019tiny}
& Stores representative samples but typically relies on appearance- or feature-based memory selection. \\

Continual semantic segmentation
& RECALL~\cite{maracani2021recall}, RECALL+~\cite{liu2023recallplus}, RCIL~\cite{zhang2022rcil}
& Develops continual semantic segmentation methods mainly for natural-image or class-incremental settings, whereas BBR-Net targets domain-shifted cardiac segmentation. \\

Domain continual medical segmentation
& Tri-enhanced distillation~\cite{zhu2024boosting}
& Uses distillation to preserve knowledge across domains, whereas BBR-Net studies boundary-guided replay selection and structural reliability. \\

Source-free/domain-adaptive medical CL
& CLMS~\cite{li2025clms}
& Addresses source-free domain adaptation and morphological knowledge transfer, whereas BBR-Net focuses on replay-buffer construction and the reliability of stored structural cues. \\

Replay and memory selection for medical segmentation
& Atypical sample selection~\cite{bera_memory_2023}
& Demonstrates the importance of memory quality, while BBR-Net defines replay priority using explicit boundary complexity and class presence. \\

Memory and class-imbalance aware CL
& Rethinking data imbalance~\cite{zhao2025rethinking}
& Highlights the importance of memory composition and class balance, motivating the class-aware component of BBR-Net. \\

Dynamic memory methods
& Dynamic memory replay~\cite{perkonigg2021dynamic}
& Adapts memory allocation across domains but does not examine structural reliability of replay samples. \\

Medical CL surveys
& Recent surveys~\cite{wu2024continual,qazi2024medicalclsurvey,kumari2025continual}
& Identify domain shift, forgetting, and memory construction as key challenges but do not empirically investigate structural consistency. \\

\hline
\textbf{This work}
& \textbf{BBR-Net}
& \textbf{Investigates how structural reliability affects replay effectiveness and introduces boundary-balanced replay for continual cardiac segmentation under domain shift.} \\
\hline
\end{tabular*}
\end{table*}

\subsection{Continual Learning for Medical Image Segmentation}

Extending CL to medical image segmentation introduces additional complexity due to domain shift, annotation variability, and structural dependencies. Unlike natural images, medical images exhibit substantial variations in intensity, resolution, and noise characteristics across datasets, often driven by differences in imaging devices and acquisition protocols \cite{guan2022domain,cheplygina2019not}. At the same time, the anatomical structures of interest remain relatively consistent. This discrepancy has been identified as a central challenge in recent surveys, which emphasize the gap between appearance variability and structural invariance in medical data~\cite{wu2024continual}.

Several studies have attempted to adapt CL techniques to medical image segmentation. Recent work has explored replay, knowledge distillation, and domain adaptation strategies to mitigate catastrophic forgetting under domain shift. For example, ~\cite{zhu2024boosting} proposed a tri-enhanced distillation framework that preserves prior knowledge through complementary distillation objectives, while  \cite{li2025clms} introduced a source-free continual segmentation approach that leverages knowledge transfer to bridge domain gaps without retaining source data. While these approaches improve robustness to varying degrees, they primarily focus on preserving feature representations, prediction consistency, or data distributions. Their emphasis on appearance-level information provides limited consideration of the anatomical structure that fundamentally defines segmentation tasks.

A closer examination of these methods reveals a recurring limitation. By emphasizing appearance similarity or feature alignment, they may struggle when domain shifts are substantial, even if the underlying anatomical structures remain unchanged. This can lead to inconsistent segmentation boundaries and degraded structural fidelity. Recent surveys further identify domain shift, annotation variability, and robust knowledge preservation as key open challenges in continual medical image analysis, highlighting the need for approaches that explicitly account for anatomical structure during continual adaptation~\cite{wu2024continual,qazi2024medicalclsurvey}.

Moreover, recent work on class-incremental surgical instrument segmentation has shown that memory composition and class imbalance substantially influence CL performance, further emphasizing the importance of informed replay sample selection beyond uniform memory construction~\cite{zhao2025rethinking}.



\subsection{Replay-Based Methods and Sample Selection}

A critical component of replay-based CL is sample selection under limited memory constraints. Standard approaches typically employ uniform sampling, which treats all samples equally regardless of their informativeness \cite{bera_memory_2023}. While simple, this strategy often fails to preserve the most relevant information for maintaining task performance \cite{aljundi2019gradient, buzzega2020dark}.

To address this, several works have explored more selective replay strategies. Hard sample mining prioritizes samples with high loss or uncertainty, under the assumption that these samples are more informative for learning \cite{aljundi2019gradient}. Although effective in improving training efficiency, such approaches remain dependent on appearance-based criteria and may neglect structurally important regions. Prototype-based methods, as used in iCaRL~\cite{rebuffi2017icarl}, compress knowledge into feature representations, offering memory efficiency at the cost of losing spatial detail. This trade-off is particularly problematic in segmentation, where spatial precision and boundary delineation are critical.

More recent efforts have incorporated distribution-aware or diversity-based sampling to improve coverage of the data space \cite{bang2021rainbow, zhu2023memory}. However, these methods still operate at the level of global feature distributions and do not explicitly account for anatomical structure. As noted in recent surveys~\cite{wu2024continual}, current replay mechanisms largely fail to capture the structural dependencies required for robust segmentation under domain shift.

\subsection{Limitations and Research Gap}
Across these categories, a common pattern emerges. Existing CL methods based on regularization, replay, or feature alignment primarily focus on preserving appearance or distributional similarity. This emphasis may be insufficient for medical image segmentation, where structural information, such as anatomical boundaries and spatial relationships, plays an important role in maintaining consistent representations across domains.

Moreover, the literature has largely overlooked the interaction between representation quality and task order. Most studies assume that tasks are presented in arbitrary order without analyzing how the initial task influences subsequent learning. Recent surveys~\cite{qazi2024medicalclsurvey} explicitly identify task ordering and data quality as underexplored factors in CL performance. The absence of systematic analysis in this regard leaves an important aspect of CL unexplored.

\subsection{Summary}

This study addresses these limitations by introducing a structural perspective on CL. Specifically, we propose a BBR-Net framework that prioritizes anatomically informative regions, thereby aligning memory selection with the geometric nature of segmentation tasks. Unlike existing approaches, which treat all samples or features equally, our method explicitly emphasizes boundary information and class balance. In addition, we provide a systematic analysis of task-order effects under domain shift, revealing that the effectiveness of replay is conditional on the quality of learned representations. This perspective extends beyond incremental performance improvements, offering new insights into the mechanisms underlying CL in medical imaging. By integrating structure-aware replay with a principled analysis of learning dynamics, this work contributes to both the theoretical understanding and practical advancement of CL for medical image segmentation.

\section{Proposed Methodology} 
\label{propose}

\begin{figure*}[!t] 
\centering 
\includegraphics[width=1\textwidth]{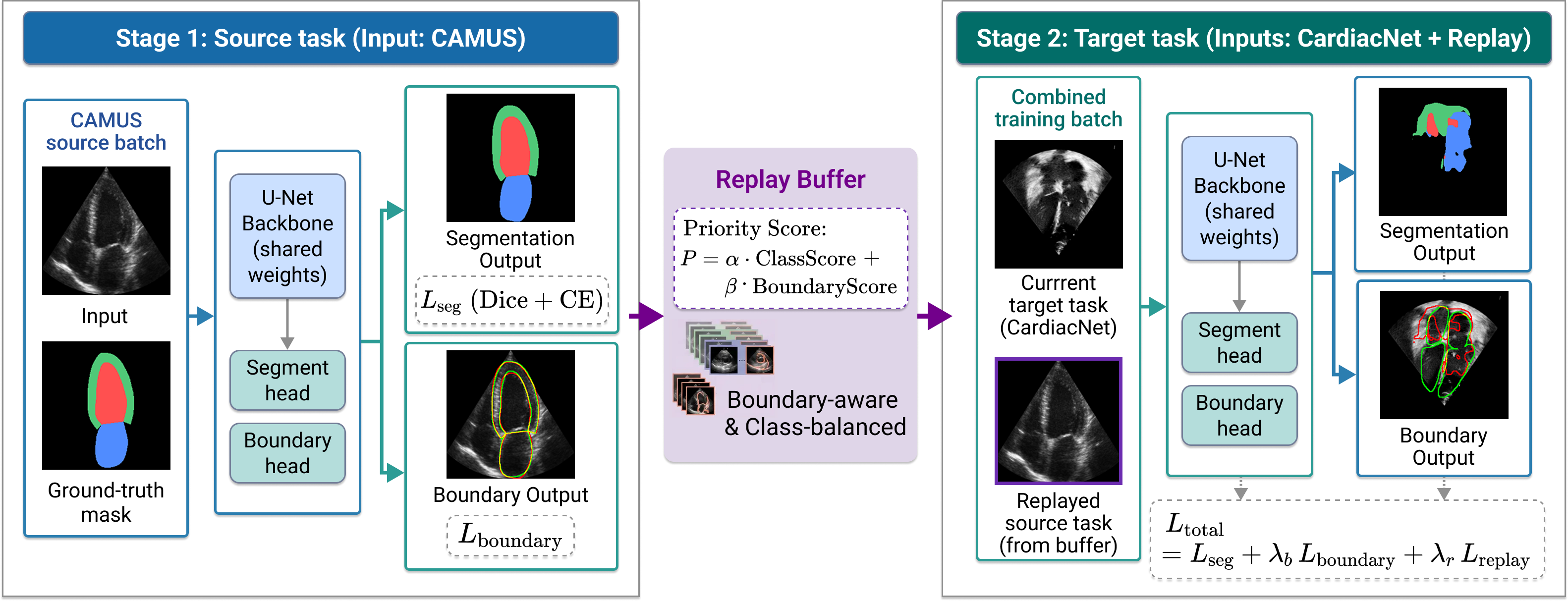} 
\caption{Training workflow of the proposed BBR-Net framework. The framework follows a two-stage CL process for medical image segmentation. In Stage 1, a dual-head U-Net is trained on the source task, CAMUS, with segmentation and boundary supervision to learn structurally consistent anatomical representations. Source-task samples are then used to construct a replay buffer according to class-balanced and boundary-aware priorities. In Stage 2, the model is adapted to the target task, CardiacNet, by jointly training on current target-task samples and replayed source-task samples. The total objective combines segmentation, boundary, and replay losses to preserve structural consistency and reduce catastrophic forgetting. \(L_{\mathrm{seg}}\), \(L_{\mathrm{boundary}}\), and \(L_{\mathrm{replay}}\) denote segmentation, boundary, and replay losses, respectively. CE denotes cross-entropy. Red, green, and blue regions denote the left ventricular endocardium, left ventricular epicardium, and left atrium, respectively, while yellow contours indicate extracted boundary maps.}
\label{fig:bbr_framework}
\label{Framework} 
\end{figure*}

\subsection{Overview}

We propose a structure-aware CL framework for multi-class cardiac image segmentation that explicitly integrates anatomical boundary information into the replay process. The key limitation of existing replay-based methods is that they treat all samples as equally informative, which is inappropriate for medical imaging tasks where anatomical structure provides more stable and transferable information than appearance. Consequently, appearance-driven replay often preserves superficial patterns while failing to retain structural consistency across tasks. To address this issue, the proposed framework prioritizes samples based on structural complexity and class distribution, ensuring that the replay buffer is composed of anatomically informative examples. The model is trained sequentially across tasks, where knowledge from the source dataset is preserved through a boundary-aware memory mechanism during adaptation to the target dataset.

As illustrated in Fig. \ref{Framework}, the learning process consists of two sequential stages. In the first stage, the model is trained on the source dataset (CAMUS) to learn stable anatomical representations. In the second stage, the model is adapted to the target dataset (CardiacNet) while selectively replaying structurally important samples stored in a memory buffer constructed from the source task. Unlike conventional replay strategies, which rely on random or heuristic selection, the proposed approach explicitly quantifies structural importance and class imbalance to guide both memory construction and sampling.

The segmentation network is formulated as a dual-head model:
\begin{equation}
f_\theta(x) = \{f_{\text{seg}}(x), f_{\text{bnd}}(x)\},
\end{equation}
where $f_{\text{seg}}$ predicts segmentation masks and $f_{\text{bnd}}$ predicts boundary maps that encode anatomical structure. The boundary branch acts as an auxiliary constraint that reinforces shape consistency during both training and replay.

\subsection{Problem Formulation}
\label{pf}

We consider a sequential learning setting with two tasks:
\begin{equation}
\mathcal{T}_1 = \text{CAMUS}, \quad \mathcal{T}_2 = \text{CardiacNet}.
\end{equation}

Each dataset is defined as:
\begin{equation}
\mathcal{D}_t = \{(x_i^{(t)}, y_i^{(t)})\}_{i=1}^{N_t},
\end{equation}
where $x_i \in \mathbb{R}^{H \times W}$ denotes an input image and $y_i \in \{0,1,\dots,C-1\}^{H \times W}$ is the corresponding segmentation mask with $C=4$ anatomical classes.

The predicted segmentation is obtained as:
\begin{equation}
\hat{y}(x) = \arg\max_c f_{\text{seg}}(x)_c.
\end{equation}

The objective is to learn model parameters $\theta$ sequentially such that performance on the first task is preserved after learning the second task. Catastrophic forgetting is quantified as:
\begin{equation}
\mathcal{F} = D^{(1)}_{\text{before}} - D^{(1)}_{\text{after}},
\end{equation}
where $D^{(1)}$ denotes the Dice score on the source task before and after learning the target task.

\subsection{Overall Training Strategy}

Training is performed in two stages. In the first stage, the model is optimized using only the source dataset:
\begin{equation}
\theta^{*}_1 = \arg\min_{\theta} \mathbb{E}_{(x,y)\sim \mathcal{D}_1} \mathcal{L}(x,y).
\end{equation}

In the second stage, the model is trained on the target dataset while incorporating replay from a memory buffer:
\begin{equation}
\theta^{*}_2 = \arg\min_{\theta} \left[
\mathbb{E}_{(x,y)\sim \mathcal{D}_2} \mathcal{L}(x,y) + 
\lambda_r \mathbb{E}_{(x,y)\sim \mathcal{M}} \mathcal{L}(x,y)
\right].
\end{equation}

Here, $\mathcal{M}$ denotes the replay buffer constructed from the source task, and $\lambda_r$ controls the contribution of replay samples. The effectiveness of the proposed method lies in how $\mathcal{M}$ is constructed and sampled.

\subsection{Boundary-Aware Replay Strategy}
\label{boundary}

\begin{figure*}[!t] 
\centering 
\includegraphics[width=1\textwidth]{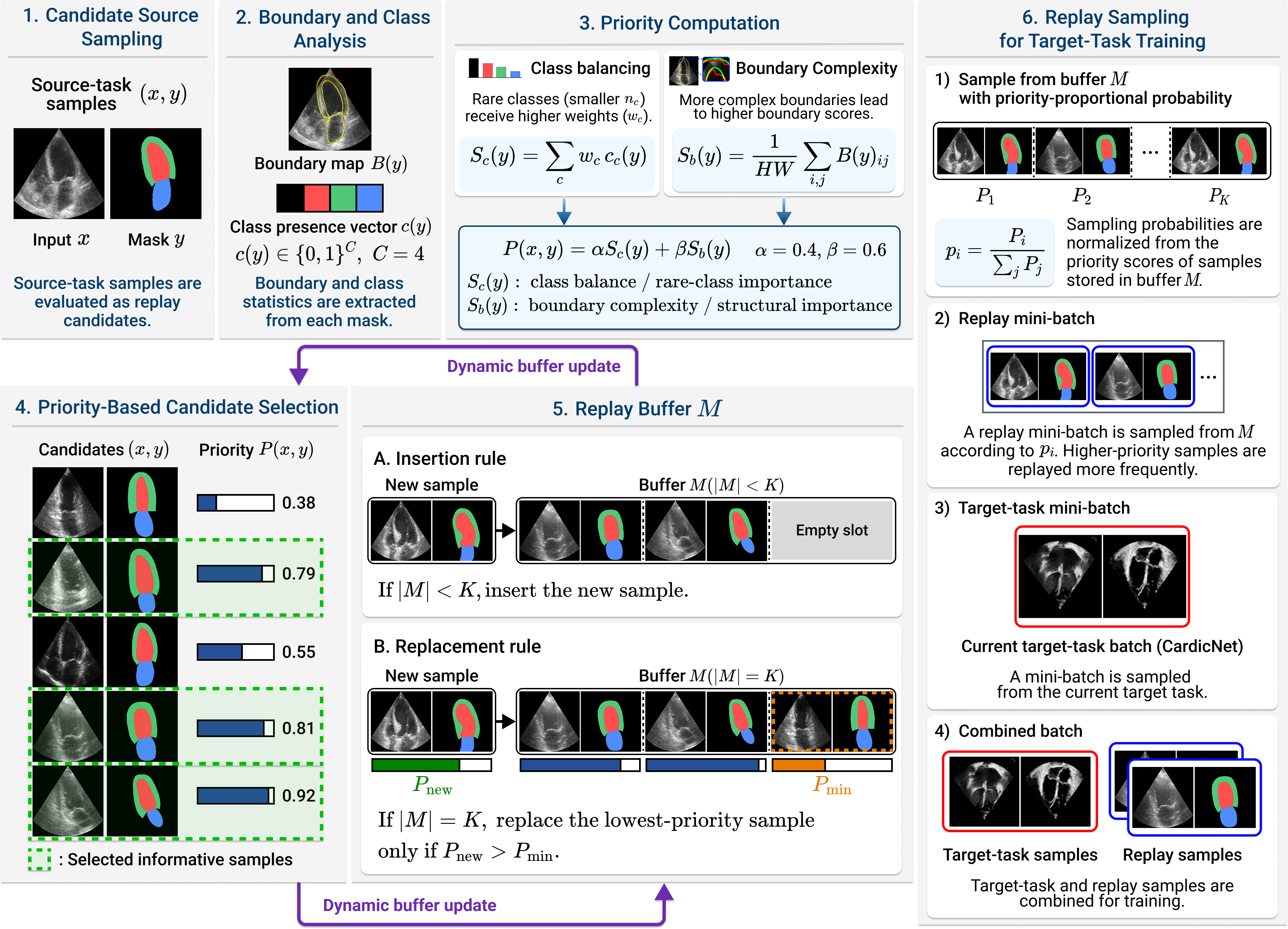} 
\caption{Overview of the proposed boundary-aware class-balanced replay mechanism. Source-task samples are evaluated as replay candidates using class-presence and boundary-complexity statistics, which are combined into a priority score for replay-buffer construction. High-priority samples are inserted or retained through priority-based buffer updates, and replay mini-batches are sampled according to priority-proportional probabilities during target-task training. Red, green, and blue regions denote the left ventricular endocardium, left ventricular epicardium, and left atrium, respectively, while yellow contours indicate extracted boundary maps.}
\label{fig2} 
\end{figure*}

The proposed approach leverages anatomical boundaries as a proxy for structural information as depicted in Fig. \ref{fig2}. Unlike pixel-wise appearance, boundaries encode shape and spatial relationships that remain more consistent across datasets.

Given a segmentation mask $y$, the boundary map is defined as:
\begin{equation}
B(y)_{ij} =
\begin{cases}
1, & \text{if } \max_{k \in \mathcal{N}(i,j)} y_k \neq \min_{k \in \mathcal{N}(i,j)} y_k, \\
0, & \text{otherwise},
\end{cases}
\end{equation}
where $\mathcal{N}(i,j)$ denotes a local neighborhood.

The structural importance of a sample is quantified as:
\begin{equation}
S_b(y) = \frac{1}{HW} \sum_{i,j} B(y)_{ij}.
\end{equation}

This score provides a simple proxy for boundary complexity and anatomical contour richness. Samples with higher structural complexity are more informative and harder to relearn, making them critical for mitigating forgetting.

\subsection{Class-Balanced Importance Modeling}
\label{class_bal}

To address class imbalance, we define a class-aware importance score. For each mask $y$, a class presence vector $\mathbf{c}(y) \in \{0,1\}^C$ indicates the presence of each class.

The frequency of each class is computed as:
\begin{equation}
n_c = \sum_i \mathbf{c}_c(y_i),
\end{equation}
and inverse weighting is applied:
\begin{equation}
w_c = \frac{1}{n_c}.
\end{equation}

The class importance score is defined as:
\begin{equation}
S_c(y) = \sum_{c=1}^{C} w_c \cdot \mathbf{c}_c(y).
\end{equation}

This formulation ensures that rare anatomical structures are prioritized during replay.

\subsection{Priority-Based Replay Buffer}
\label{pD_replay}

The replay buffer $\mathcal{M}$ maintains a fixed capacity $K$ and stores samples based on a combined priority score:
\begin{equation}
P(x,y) = \alpha S_c(y) + \beta S_b(y),
\end{equation}
where $\alpha$ and $\beta$ balance class importance and structural complexity.

When inserting a new sample, it is added if the buffer is not full. Otherwise, it replaces the sample with the lowest priority:
\begin{equation}
(x_{\min}, y_{\min}) = \arg\min_{(x,y)\in \mathcal{M}} P(x,y),
\end{equation}
if its priority is higher.

Sampling from the buffer follows a probability distribution:
\begin{equation}
p_i = \frac{P_i}{\sum_j P_j},
\end{equation}
ensuring that more informative samples are replayed more frequently.

\subsection{Choice of Backbone Architecture}

The proposed framework employs U-Net as the segmentation backbone. While more complex architectures such as HRNet \cite{wang2020deep} or transformer-based models including TransUNet and Swin-Unet \cite{chen2021transunet,cao2022swin} may provide higher absolute performance, they introduce additional architectural complexity and optimization dynamics that can obscure the analysis of CL behavior. 

U-Net \cite{ronneberger2015unet} offers a well-balanced architecture with an encoder–decoder structure and skip connections that preserve both global context and fine-grained spatial information. Its stable training behavior and widespread adoption in medical imaging make it a suitable choice for isolating the effect of the proposed replay mechanism. Importantly, the use of U-Net ensures that performance improvements can be attributed to the structure-aware replay strategy rather than architectural complexity. Furthermore, the multi-scale feature fusion inherent in U-Net naturally supports boundary prediction, enabling effective integration of the auxiliary boundary branch without requiring architectural modifications.

\subsection{Loss Function}

The proposed framework is optimized using a multi-task objective that jointly enforces semantic segmentation accuracy and structural consistency. The overall loss consists of three components: segmentation loss, boundary loss, and replay loss.

The segmentation loss combines cross-entropy and Dice loss to balance pixel-wise classification and region-level overlap:
\begin{equation}
\mathcal{L}_{seg} = \mathcal{L}_{CE} + \mathcal{L}_{Dice},
\end{equation}
where the cross-entropy loss is defined as:
\begin{equation}
\mathcal{L}_{CE} = -\sum_{c=1}^{C} y_c \log p_c,
\end{equation}
and the Dice loss is given by:
\begin{equation}
\mathcal{L}_{Dice} = 1 - \frac{2 \sum p \cdot y}{\sum p + \sum y + \epsilon}.
\end{equation}
Here, $p$ denotes the predicted probability map and $y$ represents the ground truth mask.

To explicitly enforce structural consistency, a boundary loss is introduced using binary cross-entropy (BCE):
\begin{equation}
\mathcal{L}_{bnd} = \text{BCE}(f_{\text{bnd}}(x), B(y)),
\end{equation}
where $B(y)$ denotes the ground truth boundary map. This term encourages accurate localization of anatomical contours.

During CL, replay samples are incorporated into training. The replay loss is defined using the same objective applied to samples drawn from the memory buffer:
\begin{equation}
\mathcal{L}_{replay} = \mathcal{L}_{seg}(x_r, y_r) + \lambda_b \mathcal{L}_{bnd}(x_r, y_r).
\end{equation}

The final training objective is:
\begin{equation}
\mathcal{L} =
\mathcal{L}_{seg}
+ \lambda_b \mathcal{L}_{bnd}
+ \lambda_r \mathcal{L}_{replay},
\end{equation}
where $\lambda_b$ and $\lambda_r$ control the contributions of boundary supervision and replay, respectively.

Unlike conventional replay methods that treat all samples equally, the proposed method explicitly models anatomical structure. This shifts the learning paradigm from appearance preservation to structure preservation, which is critical for medical image segmentation under domain shift. However, as demonstrated in later sections, this assumption holds strongly only when structural distributions are consistent across tasks.

\begin{algorithm}[t]
\caption{Boundary-Aware Class-Balanced Replay for Continual Segmentation}
\label{alg:bbr}
\footnotesize
\begin{algorithmic}[1]

\Require Source dataset $\mathcal{D}_1$, target dataset $\mathcal{D}_2$, memory capacity $K$, replay weight $\lambda_r$, priority weights $\alpha, \beta$
\Ensure Trained model parameters $\theta^{*}$

\State Initialize model parameters $\theta$
\State Initialize empty memory buffer $\mathcal{M} \leftarrow \emptyset$

\State Define task loss $\mathcal{L}_{\mathrm{task}}(x,y)
= \mathcal{L}_{\mathrm{seg}}(x,y) + \lambda_b \mathcal{L}_{\mathrm{bnd}}(x,y)$

\Statex \textbf{Stage 1: Train on Source Task (CAMUS)}
\For{each epoch}
    \For{each mini-batch $(x, y) \sim \mathcal{D}_1$}
        \State Compute segmentation prediction $\hat{y} = f_{\mathrm{seg}}(x)$
        \State Compute boundary prediction $\hat{b} = f_{\mathrm{bnd}}(x)$
        \State Compute task loss $\mathcal{L}_{\mathrm{task}}(x,y)$
        \State Update $\theta$ using gradient descent
    \EndFor
\EndFor

\Statex \textbf{Construct Replay Buffer (Fig.~\ref{fig2}, Block 1--5)}
\For{each sample $(x,y) \in \mathcal{D}_1$}
    \State Compute boundary map $B(y)$
    \State Compute boundary score $S_b(y)$
    \State Compute class score $S_c(y)$
    \State Compute priority $P(x,y) = \alpha S_c(y) + \beta S_b(y)$
    \If{$|\mathcal{M}| < K$}
        \State Insert $(x,y)$ into $\mathcal{M}$
    \Else
        \State Find $(x_{\min}, y_{\min}) = \arg\min_{(x,y)\in \mathcal{M}} P(x,y)$
        \If{$P(x,y) > P(x_{\min}, y_{\min})$}
            \State Replace $(x_{\min}, y_{\min})$ with $(x,y)$
        \EndIf
    \EndIf
\EndFor

\Statex \textbf{Stage 2: CL with Replay (Fig.~\ref{fig2}, Block 6)}
\For{each epoch}
    \For{each mini-batch $(x, y) \sim \mathcal{D}_2$}
        \State Sample replay batch $(x_r, y_r) \sim \mathcal{M}$ using $p_i \propto P_i$
        \State Compute predictions $\hat{y} = f_{\mathrm{seg}}(x)$ and $\hat{y}_r = f_{\mathrm{seg}}(x_r)$
        \State Compute boundary predictions $\hat{b} = f_{\mathrm{bnd}}(x)$ and $\hat{b}_r = f_{\mathrm{bnd}}(x_r)$
        \State Compute total loss 
$\mathcal{L}_{\mathrm{total}} =
\mathcal{L}_{\mathrm{task}}(x,y)
+ \lambda_r \mathcal{L}_{\mathrm{task}}(x_r,y_r)$
        \State Update $\theta$ using gradient descent
    \EndFor
\EndFor

\State \Return $\theta^{*}$

\end{algorithmic}
\end{algorithm}

\subsection{Implementation Details}

This section provides the implementation details corresponding to the proposed boundary-aware replay framework described in Section \ref{propose}. The goal is to ensure reproducibility while maintaining consistency with the methodological design.

We adopt a U-Net-based architecture with an auxiliary boundary prediction head, consistent with the dual-branch formulation introduced in Section~3.1. The encoder–decoder structure consists of four downsampling and four upsampling stages with skip connections that preserve spatial information across scales. Each block contains two convolutional layers followed by batch normalization and ReLU activation.

Given an input image $x \in \mathbb{R}^{1 \times H \times W}$, the network produces:
\begin{equation}
f_\theta(x) = \{f_{\text{seg}}(x), f_{\text{bnd}}(x)\},
\end{equation}
where $f_{\text{seg}}(x) \in \mathbb{R}^{C \times H \times W}$ outputs segmentation logits and $f_{\text{bnd}}(x) \in \mathbb{R}^{1 \times H \times W}$ predicts boundary maps. The number of classes is fixed to $C=4$, corresponding to background, left ventricle endocardium, left ventricle epicardium, and left atrium.

All models are implemented in PyTorch and trained on a single GPU. The training and replay hyperparameters are summarized in Table~\ref{tab:unified_config}. The same configuration is used across all methods to ensure fair comparison. During CL, each mini-batch sampled from the current task is augmented with replay samples drawn from the memory buffer according to the priority distribution defined in Section~\ref{pD_replay}. Specifically, for each batch from $\mathcal{D}_2$, a replay batch is sampled from $\mathcal{M}$ and incorporated into the loss via the replay term weighted by $\lambda_r$.

The memory buffer has a fixed capacity of $800$ samples and is constructed using the proposed priority function that combines class-aware and boundary-aware scores. The weighting coefficients $\alpha$ and $\beta$ control the contribution of class imbalance and structural complexity, respectively.

Algorithm~\ref{alg:bbr} formalizes the replay-buffer construction, update, and sampling process illustrated in Fig.~\ref{fig2}. The boundary/class analysis and priority computation in Fig.~\ref{fig2} correspond to Lines 12--16, the priority-based buffer insertion and replacement steps correspond to Lines 17--25, and the replay sampling for target-task training corresponds to Lines 28--31.

\begin{table}[t]
\centering
\caption{Unified training and replay configuration.}
\label{tab:unified_config}
\scriptsize
\setlength{\tabcolsep}{3.5pt}
\renewcommand{\arraystretch}{1.12}

\begin{tabularx}{\columnwidth}{@{}
>{\raggedright\arraybackslash}p{0.20\columnwidth}
>{\raggedright\arraybackslash}X
>{\raggedright\arraybackslash}p{0.23\columnwidth}
@{}}
\toprule
\textbf{Category} & \textbf{Parameter} & \textbf{Value} \\
\midrule

\multirow{9}{0.20\columnwidth}{\centering Training\\Settings}
& Input resolution & $256 \times 256$ \\
& Batch size & $8$ \\
& Replay batch size & $4$ \\
& Optimizer & Adam \\
& Learning rate & $1 \times 10^{-4}$ \\
& Weight decay & $1 \times 10^{-5}$ \\
& Epochs per task & $20$ \\
& Early stopping patience & $8$ \\
& Mixed precision & Enabled \\

\midrule

\multirow{6}{0.20\columnwidth}{\centering Replay\\Configuration}
& Memory capacity & $800$ samples \\
& Replay loss weight & $\lambda_r = 0.7$ \\
& Boundary loss weight & $\lambda_b = 0.3$ \\
& Class priority weight & $\alpha = 0.4$ \\
& Boundary priority weight & $\beta = 0.6$ \\
& Priority scaling exponent & $1.0$ \\

\bottomrule
\end{tabularx}
\end{table}
\FloatBarrier

To address class imbalance, inverse-frequency class weights are computed as:
\begin{equation}
w_c = \frac{1}{n_c},
\end{equation}
and incorporated into the cross-entropy loss. In addition, a weighted sampling strategy is used during training to bias mini-batch construction toward rare classes and structurally informative samples, consistent with the class-aware importance defined in Section~\ref{class_bal}.

We evaluate the proposed framework on two echocardiographic datasets used in a sequential learning setting. The CAMUS dataset is used as the source task and provides relatively clean and consistent annotations, enabling the model to learn stable anatomical representations. The CardiacNet dataset is used as the target task and contains more diverse and noisy clinical samples, including atrial septal defect (ASD) and non-ASD cases, making it suitable for evaluating CL under domain shift.

All volumetric data are converted into 2D slices. Each slice is normalized to the range $[0,1]$ and resized to $256 \times 256$. Boundary maps are computed on-the-fly from segmentation masks using the formulation described in Section~\ref{boundary}.

To ensure consistency across datasets, all labels are unified into four classes:
0: background, 1: left ventricular (LV) endocardium, 2: LV epicardium, 3: left atrium.
For CardiacNet, label $4$ is remapped to class $3$ to maintain semantic alignment between datasets. Datasets are split at the patient level into training and validation sets to avoid data leakage. For CAMUS, predefined splits are used when available; otherwise, a random split with a validation ratio of $20\%$ is applied. CardiacNet is split randomly at the patient level with the same ratio. All experiments are conducted using three random seeds $\{42, 123, 999\}$ to ensure robustness. Final results are reported as the mean and standard deviation across seeds.

In summary, the entire pipeline, including preprocessing, boundary computation, replay buffer construction, and training, is implemented within a unified framework. The configuration is fixed across all experiments to ensure that performance differences arise solely from methodological variations rather than implementation inconsistencies.

\begin{table}[t]
\centering
\caption{Experimental protocol and evaluation setup.}
\label{tab:protocol}
\footnotesize
\setlength{\tabcolsep}{4pt}
\renewcommand{\arraystretch}{1.12}

\begin{tabularx}{\columnwidth}{@{}
>{\raggedright\arraybackslash}p{0.39\columnwidth}
>{\raggedright\arraybackslash}X
@{}}
\toprule
\textbf{Aspect} & \textbf{Details} \\
\midrule
Forward task sequence & CAMUS to CardiacNet \\
Reverse task sequence & CardiacNet to CAMUS \\
Learning setting & Sequential CL \\
Replay usage & Applied only during the second task \\
Data split strategy & Patient-wise split \\
Validation ratio & $20\%$ \\
Number of tasks & $2$ \\
Number of classes & $4$ \\
Evaluation metrics & Dice, per-class Dice, and forgetting \\
Number of seeds & $3$ \\
Seed values & $\{42, 123, 999\}$ \\
Model selection & Best validation Dice \\
\bottomrule
\end{tabularx}
\end{table}


\section{Experimental Setup}
\label{experimental}

This section evaluates the proposed BBR-Net framework introduced in Section~3. The experiments are designed to validate the central hypothesis of this work: structural consistency, rather than appearance similarity, governs knowledge retention in continual medical image segmentation. To this end, we analyze performance under both forward and reverse task orders, allowing us to explicitly examine when replay succeeds and when it fails. Table \ref{tab:protocol} summarizes the experimental protocol and evaluation settings.

\subsection{Datasets}

We consider a two-task CL setting consistent with the formulation in Section~\ref{pf}.

\subsubsection{CAMUS Dataset}  
The CAMUS dataset \cite{camus} is used as the first task in the forward setting. It provides high-quality annotations for cardiac structures, including the left ventricle endocardium, left ventricle epicardium, and left atrium. Due to its clean and anatomically consistent nature, it enables the model to learn stable structural representations, which is critical for the proposed boundary-aware replay mechanism. A summary of the dataset statistics is provided in
Table~\ref{tab:dataset_stats}.
\begin{table}[!t]
\centering
\caption{Dataset statistics used in the continual learning experiments.}
\label{tab:dataset_stats}
\footnotesize
\renewcommand{\arraystretch}{1.12}
\setlength{\tabcolsep}{5pt}
\begin{tabular*}{\linewidth}{@{\extracolsep{\fill}}lcc@{}}
\hline
\textbf{Dataset} 
& \textbf{Patients (train/val.)} 
& \textbf{Slices (train/val.)} \\
\hline
CAMUS      & 400 / 100 & 17,006 / 4,226 \\
CardiacNet & 110 / 27  & 9,892 / 2,135 \\
\hline
\end{tabular*}
\end{table}
\subsubsection{CardiacNet Dataset}  
The CardiacNet dataset \cite{yangcardiacnet} is used as the second task in the forward setting. It contains data collected from diverse clinical environments, including ASD and non-ASD cases. Compared to CAMUS, it exhibits higher variability, noise, and structural ambiguity. This makes it a challenging benchmark for evaluating whether the replay mechanism can preserve previously learned structural knowledge under domain shift.


\subsection{Evaluation Metrics}

The evaluation protocol directly follows the objectives defined in Section~\ref{pf}. The training process is explained in Algorithm \ref{alg:bbr}.


\subsubsection{Dice Score}  
The Dice similarity coefficient is used as the primary metric to evaluate segmentation accuracy. For a predicted segmentation $p$ and ground truth $y$, the Dice score is defined as:
\begin{equation}
\text{Dice}(p, y) = \frac{2 \sum_{i} p_i y_i}{\sum_{i} p_i + \sum_{i} y_i},
\end{equation}
where $p_i \in [0,1]$ denotes the predicted probability for pixel $i$, and $y_i \in \{0,1\}$ is the corresponding ground truth label.

\subsubsection{Per-Class Dice}  
To evaluate performance across different anatomical structures, we compute the Dice score for each class separately. For class $c \in \{1, \dots, C\}$, the per-class Dice is given by:
\begin{equation}
\text{Dice}_c = \frac{2 \sum_{i} p_{i,c} y_{i,c}}{\sum_{i} p_{i,c} + \sum_{i} y_{i,c}},
\end{equation}
where $p_{i,c}$ and $y_{i,c}$ denote the predicted and ground truth values for class $c$ at pixel $i$. The overall Dice score is obtained by averaging over all classes:
\begin{equation}
\text{Dice}_{\text{mean}} = \frac{1}{C} \sum_{c=1}^{C} \text{Dice}_c.
\end{equation}

\subsubsection{Forgetting Measure}  
To quantify catastrophic forgetting, we compute the difference between performance before and after learning the subsequent task:
\begin{equation}
\mathcal{F} = D_{\text{before}} - D_{\text{after}},
\end{equation}
where $D_{\text{before}}$ denotes the Dice score on the first task after initial training, and $D_{\text{after}}$ denotes the Dice score on the same task after learning the second task. A higher value of $\mathcal{F}$ indicates more severe forgetting, while a value close to zero indicates effective knowledge retention. All methods are evaluated under identical settings to ensure that performance differences arise solely from the replay strategy rather than implementation variations.

\section{Results}
\label{results}

This section evaluates the proposed BBR-Net framework with respect to three key objectives: (1) mitigating catastrophic forgetting, (2) preserving structural consistency, and (3) understanding the role of task order under domain shift.
\begin{table*}[t]
\centering
\caption{Results under forward and reverse task orders. All methods are evaluated over three random seeds and reported as mean $\pm$ standard deviation.}
\label{tab:forward_reverse_results}
\footnotesize
\setlength{\tabcolsep}{5pt}
\renewcommand{\arraystretch}{1.12}

\begin{threeparttable}

\begin{tabular*}{\textwidth}{@{\extracolsep{\fill}}lccc@{}}
\toprule
\multicolumn{4}{@{}l@{}}{\textbf{Forward order: CAMUS $\rightarrow$ CardiacNet}} \\
\midrule
\textbf{Method}
& \makecell{\textbf{CAMUS}\\\textbf{(After) $\uparrow$}}
& \makecell{\textbf{CardiacNet}\\\textbf{$\uparrow$}}
& \makecell{\textbf{Forgetting}\\\textbf{$\downarrow$}} \\
\midrule

Fine-tuning 
& $0.320 \pm 0.043$ 
& $0.795 \pm 0.011$ 
& $0.581 \pm 0.043$ \\

Replay 
& $0.882 \pm 0.006$ 
& $0.746 \pm 0.008$ 
& $0.019 \pm 0.007$ \\

Weighted Sampler Replay 
& $0.892 \pm 0.002$ 
& $0.752 \pm 0.003$ 
& $0.009 \pm 0.002$ \\

Boundary-aware Replay 
& $0.883 \pm 0.009$ 
& $0.748 \pm 0.012$ 
& $0.018 \pm 0.009$ \\

BBR-Net (Proposed) 
& $0.893 \pm 0.002$ 
& $0.777 \pm 0.007$ 
& $0.009 \pm 0.003$ \\

Joint Training (Reference) 
& $0.902 \pm 0.001$ 
& $0.771 \pm 0.008$ 
& -- \\

\midrule[\heavyrulewidth]

\multicolumn{4}{@{}l@{}}{\textbf{Reverse order: CardiacNet $\rightarrow$ CAMUS}} \\
\midrule
\textbf{Method}
& \makecell{\textbf{CardiacNet}\\\textbf{(After) $\uparrow$}}
& \makecell{\textbf{CAMUS}\\\textbf{$\uparrow$}}
& \makecell{\textbf{Forgetting}\\\textbf{$\downarrow$}} \\
\midrule

Fine-tuning 
& $0.181 \pm 0.003$ 
& $0.902 \pm 0.001$ 
& $0.610 \pm 0.010$ \\

Replay 
& $0.759 \pm 0.008$ 
& $0.901 \pm 0.001$ 
& $0.040 \pm 0.010$ \\

Weighted Sampler Replay 
& $0.786 \pm 0.004$ 
& $0.900 \pm 0.000$ 
& $\mathbf{0.015} \pm 0.004$ \\

Boundary-aware Replay 
& $0.197 \pm 0.052$ 
& $0.901 \pm 0.001$ 
& $0.594 \pm 0.064$ \\

BBR-Net (Proposed)
& $0.198 \pm 0.063$ 
& $0.899 \pm 0.001$ 
& $0.595 \pm 0.066$ \\

\bottomrule
\end{tabular*}

\begin{tablenotes}
\footnotesize
\item Joint training is reported as an offline full-data reference and is not a continual-learning method; forgetting is therefore not reported.
\end{tablenotes}

\end{threeparttable}
\end{table*}
\subsection{Baselines}

We compare the proposed method against representative baselines that isolate different aspects of CL. Fine-tuning represents sequential learning without memory and exposes the extent of catastrophic forgetting. Plain replay evaluates the benefit of memory without structural prioritization. Joint training is included as an offline full-data reference. This setup enables a controlled analysis of how structural replay contributes beyond standard approaches.

\subsection{Mitigating Catastrophic Forgetting}

Table~\ref{tab:forward_reverse_results} shows that fine-tuning suffers from severe forgetting, with CAMUS performance dropping to $0.320$. This confirms that, in the absence of memory, previously learned knowledge is almost entirely overwritten. Introducing replay improves retention significantly, increasing CAMUS performance to $0.882$. However, the remaining gap indicates that standard replay substantially improves retention, but BBR-Net provides the best overall balance between source-task retention and target-task adaptation. In contrast, the proposed method achieves $0.893$ Dice with a forgetting score of $0.009$, close to the offline joint-training reference ($0.902$). This demonstrates that the model retains almost all previously learned information. These results indicate that catastrophic forgetting in medical segmentation is not merely a consequence of sequential training, but is strongly influenced by how past information is selected and preserved. Structure-aware replay provides a more reliable mechanism for long-term retention than uniform replay.

\subsection{Structural Consistency and Representation Quality}

A key difference between the proposed method and standard replay lies in how replay samples are selected. While plain replay treats all samples equally, the proposed method prioritizes boundary regions and maintains class balance.

This design leads to consistent improvements in segmentation quality, particularly in preserving anatomical structures. The performance gain from $0.882$ (replay) to $0.893$ (proposed) cannot be attributed solely to optimization, as both methods use the same training protocol. Instead, it reflects improved representation quality driven by structurally informative samples. These findings suggest that preserving structural information, rather than global appearance is critical for maintaining segmentation performance across tasks. Boundary-aware sampling acts as an implicit regularizer, stabilizing spatial representations and preventing structural drift.

\subsection{Stability–Plasticity Trade-off}

While the proposed method significantly improves retention, it achieves slightly lower performance on the new task (CardiacNet: $0.777$) compared to fine-tuning ($0.795$). This reflects a controlled trade-off between stability and plasticity. Unlike fine-tuning, which aggressively adapts to the new domain at the cost of forgetting, the proposed method constrains updates to preserve prior knowledge. The small reduction in new-task performance is therefore expected and indicates balanced learning. In CL, optimal performance is not defined by maximizing accuracy on the current task, but by maintaining performance across all tasks. The proposed method achieves this balance effectively, prioritizing long-term knowledge retention over short-term gains.

\subsection{Task-Order Sensitivity and Domain Shift}

The effectiveness of replay is further examined under different task orders. In the forward setting (CAMUS to CardiacNet), the model learns from clean and structurally consistent data first, leading to stable representations and effective replay. In contrast, when training begins with noisier and more heterogeneous data, the quality of learned representations degrades. As a result, replay becomes less effective, and forgetting increases significantly. These observations highlight that replay-based CL is highly dependent on the quality of initial representations. Structural consistency in early training stages plays a critical role in determining whether replay will succeed or fail.



\subsection{Overall Findings}

Across all experiments, a consistent pattern emerges regarding the role of structural reliability in replay-based CL. In the forward task order, both weighted replay and the proposed BBR-Net achieve strong source-task retention and low forgetting, indicating that replay can be highly effective when stable anatomical priors are learned early. However, the proposed method provides the best overall balance between retention and target-task adaptation while explicitly incorporating structural information into replay selection.

The controlled structural perturbation analysis further reveals that replay effectiveness depends strongly on the structural quality of stored samples. As boundary corruption increases, forgetting rises consistently despite identical training settings and memory capacity. This observation suggests that replay performance is influenced not only by sample frequency or distribution coverage, but also by the anatomical reliability of replay representations.

More importantly, the reverse task-order experiments demonstrate that structure-aware replay mechanisms become less effective when the initial structural priors are unreliable. Under this condition, boundary-aware replay may unintentionally amplify unreliable structural information, leading to severe forgetting despite the presence of memory. These findings support the central hypothesis of this work: structural consistency is an important factor in continual medical image segmentation, and replay mechanisms that ignore structural reliability may remain limited under strong domain shifts. Across the three random seeds, BBR-Net consistently reduced forgetting compared with standard Replay while maintaining higher source-task retention. Although Weighted Sampler Replay achieved competitive retention in the forward task order, BBR-Net provided a more favorable balance between source-task preservation and target-task adaptation by combining class-aware sampling with boundary-aware structural prioritization.

\subsection{Reverse Task Order Analysis (CardiacNet to CAMUS)}

The reverse task order results in Table~\ref{tab:forward_reverse_results} reveal a critical and non-trivial behavior of replay-based CL under structural domain shift. In contrast to the forward setting, where structure-aware replay is highly effective, the reverse setting exhibits a fundamentally different trend. First, standard replay significantly improves retention compared to fine-tuning, reducing forgetting from $0.610$ to $0.040$. Furthermore, weighted sampler replay achieves the best overall performance, with the lowest forgetting ($0.015$) and highest retention on CardiacNet. This indicates that replay mechanisms remain effective even under reverse task order, provided that they do not rely heavily on structural assumptions.

However, in sharp contrast, both boundary-aware replay and the full BBR-Net fail to preserve prior knowledge, exhibiting forgetting levels ($\approx 0.595$) comparable to fine-tuning. This behavior is particularly important, as it demonstrates that the proposed structure-aware prioritization mechanism is not universally beneficial. This degradation can be attributed to the nature of the initial task. When training begins on CardiacNet, the model learns representations from data that are inherently noisy and structurally inconsistent due to variability in acquisition conditions and annotation quality. Consequently, the boundary information extracted at this stage is unreliable. When such noisy structural cues are used to guide replay selection, the memory buffer becomes biased toward suboptimal or misleading samples.

During subsequent training on CAMUS, these structurally inconsistent replay samples fail to reinforce correct anatomical relationships and instead propagate errors, leading to severe forgetting. In contrast, weighted sampler replay, which does not depend on structural priors, avoids this issue and maintains stable performance. These findings highlight a key insight: the effectiveness of structure-aware replay is conditional on the quality of the learned structural priors. When stable anatomical structure is learned early (forward order), boundary-aware replay is highly effective. However, when initial representations are noisy (reverse order), structure-based prioritization becomes unreliable and can even degrade performance.

An important observation is that Weighted Sampler Replay achieves retention performance comparable to BBR-Net in the forward task order and substantially outperforms it in the reverse task order. This indicates that structure-aware replay is not universally superior to simpler replay strategies. Rather, its effectiveness depends on the availability of reliable structural priors.

When stable anatomical representations are learned early, boundary-aware prioritization improves the balance between source-task retention and target-task adaptation. However, when the initial task contains noisy or structurally inconsistent annotations, emphasizing boundary information can amplify unreliable structural cues and degrade replay quality. In contrast, Weighted Sampler Replay remains less sensitive to structural noise because it does not explicitly rely on boundary-based prioritization.

Therefore, the primary contribution of BBR-Net is not that it universally outperforms all replay methods, but that it reveals when and why structure-aware replay succeeds or fails under domain shift. These findings suggest that replay effectiveness depends not only on memory construction, but also on the structural reliability of the representations used to guide sample selection.

The reverse task-order analysis revealed a consistent degradation of structure-aware replay methods under noisy initial representations. In contrast to standard Replay and Weighted Sampler Replay, both Boundary-aware Replay and Full BBR-Net exhibited substantially higher forgetting, supporting the interpretation that unreliable structural priors can negatively affect replay selection.

Overall, the reverse task order results demonstrate that CL performance is not only method-dependent but also strongly influenced by the order of tasks and the structural quality of the initial dataset. This exposes a fundamental limitation of existing replay strategies and emphasizes the need for more robust structure-aware mechanisms that can adapt to varying data quality.





\subsection{Controlled Structural Perturbation Analysis}

\label{sec:controlled_perturbation}

To directly examine whether structural reliability affects replay-based retention, we conducted a controlled structural perturbation experiment. This experiment was performed in the forward task order (CAMUS to CardiacNet) using the same dataset split, U-Net backbone, training schedule, replay memory size, and optimization settings as the main experiments. The key difference was that the structural quality of the source-task masks was deliberately degraded before constructing the replay buffer. The corruption was applied only to the source-task masks used for replay-buffer construction. The original CAMUS training masks used during source-task learning were kept unchanged, and the CardiacNet target-task labels were also kept unchanged. Thus, the perturbation did not modify the main training labels or the target-task supervision. It affected only the replay samples stored in memory after source-task learning. This design isolates the effect of structural reliability in replay memory while avoiding confounding changes in model training, dataset composition, or task identity.

We considered three corruption severity levels: 0, 3, and 5. Severity 0 denotes the clean condition, where replay masks are stored without perturbation. Severity 3 denotes moderate structural perturbation, where foreground anatomical regions are randomly shifted, eroded, or dilated within a limited range. Severity 5 denotes stronger perturbation, where the same operations are applied with larger spatial deformation. These operations alter the boundary geometry and contour reliability of the replay masks while preserving the same label space. Accordingly, higher severity indicates lower structural reliability of the stored replay masks. 

Table~\ref{tab:structural_corruption} reports the results for the controlled replay setting. This experiment is used as a diagnostic analysis rather than as the main BBR-Net comparison reported in Table~\ref{tab:forward_reverse_results}. For this reason, the clean condition in Table~\ref{tab:structural_corruption} is not intended to exactly reproduce the full BBR-Net result in Table~\ref{tab:forward_reverse_results}. The purpose of this experiment is not to establish a new best-performing model, but to test whether replay performance changes systematically when only the structural reliability of replay masks is degraded.

\begin{table}[t]
\centering
\caption{Effect of structural perturbation on replay-buffer construction in the forward task order, CAMUS to CardiacNet.}
\label{tab:structural_corruption}
\scriptsize
\setlength{\tabcolsep}{3.5pt}
\renewcommand{\arraystretch}{1.12}

\begin{threeparttable}
\begin{tabular*}{\columnwidth}{@{\extracolsep{\fill}}lccc@{}}
\toprule
\textbf{Severity}
& \makecell{\textbf{CAMUS}\\\textbf{(After) $\uparrow$}}
& \makecell{\textbf{CardiacNet}\\\textbf{$\uparrow$}}
& \makecell{\textbf{Forgetting}\\\textbf{$\downarrow$}} \\
\midrule
0 (Clean)    
& $0.890 \pm 0.004$ 
& $0.748 \pm 0.001$ 
& $0.011 \pm 0.004$ \\

3 (Moderate) 
& $0.875 \pm 0.000$ 
& $0.750 \pm 0.000$ 
& $0.025 \pm 0.001$ \\

5 (Severe)   
& $0.848 \pm 0.008$ 
& $0.758 \pm 0.004$ 
& $0.052 \pm 0.009$ \\
\bottomrule
\end{tabular*}

\begin{tablenotes}
\scriptsize
\item Corruption is applied only to source-task masks used for replay-buffer construction. Source-task training labels and target-task labels remain unchanged. Severity 0 denotes clean replay masks, while severities 3 and 5 indicate moderate and severe boundary perturbations, respectively.
\end{tablenotes}
\end{threeparttable}
\end{table}

\begin{figure*}[t]
\centering

\begin{subfigure}[t]{0.49\textwidth}
\centering
\begin{tikzpicture}
\begin{axis}[
    width=\linewidth,
    height=0.64\linewidth,
    xlabel={Replay-mask corruption severity},
    ylabel={Dice score},
    xtick={0,3,5},
    xticklabels={0,3,5},
    ymin=0.70, ymax=0.92,
    grid=major,
    grid style={dashed, gray!30},
    legend style={
        at={(0.5,1.03)},
        anchor=south,
        legend columns=2,
        draw=none,
        fill=none,
        font=\scriptsize,
        /tikz/every even column/.append style={column sep=0.4cm}
    },
    legend cell align={left},
    tick label style={font=\scriptsize},
    label style={font=\scriptsize},
    line width=0.8pt
]
\addplot[
    mark=*,
    mark size=2pt
] coordinates {
    (0,0.890)
    (3,0.875)
    (5,0.848)
};

\addplot[
    mark=square*,
    mark size=2pt,
    dashed
] coordinates {
    (0,0.748)
    (3,0.750)
    (5,0.758)
};

\legend{CAMUS retention, CardiacNet adaptation}
\end{axis}
\end{tikzpicture}
\caption{Dice scores under increasing structural corruption.}
\label{fig:corruption_tradeoff}
\end{subfigure}
\hfill
\begin{subfigure}[t]{0.49\textwidth}
\centering
\begin{tikzpicture}
\begin{axis}[
    width=\linewidth,
    height=0.64\linewidth,
    xlabel={Replay-mask corruption severity},
    ylabel={Forgetting score},
    xtick={0,3,5},
    xticklabels={0,3,5},
    ymin=0, ymax=0.06,
    ytick={0,0.02,0.04,0.06},
    yticklabels={0.00,0.02,0.04,0.06},
    scaled y ticks=false,
    grid=major,
    grid style={dashed, gray!30},
    tick label style={font=\scriptsize},
    label style={font=\scriptsize},
    line width=0.8pt
]
\addplot[
    mark=*,
    mark size=2pt
] coordinates {
    (0,0.011)
    (3,0.025)
    (5,0.052)
};
\end{axis}
\end{tikzpicture}
\caption{Forgetting increases with structural corruption.}
\label{fig:corruption_forgetting}
\end{subfigure}

\caption{Structural corruption analysis for the forward task order (CAMUS to CardiacNet). As source-task replay masks are progressively corrupted, CAMUS retention decreases while CardiacNet performance remains stable or slightly improves. Forgetting increases with corruption severity, indicating that replay effectiveness depends on the structural reliability of stored samples.}
\label{fig:structural_corruption_analysis}
\end{figure*}
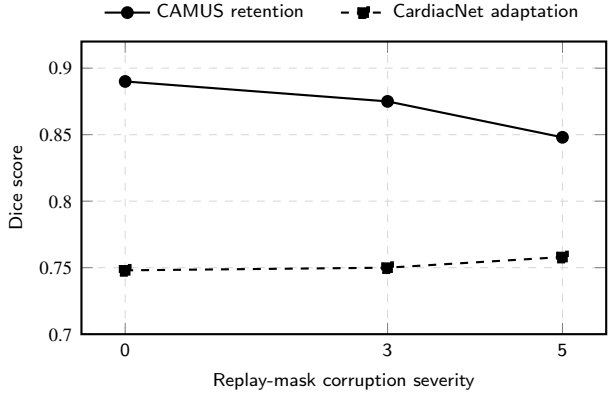
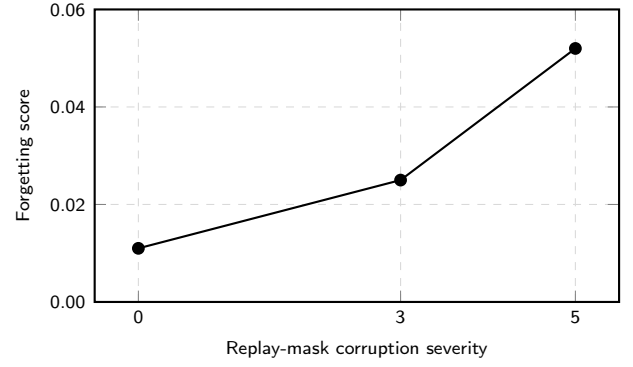

As corruption severity increases, CAMUS retention after learning CardiacNet decreases from $0.890$ under clean replay to $0.848$ under severe corruption. At the same time, forgetting increases from $0.011$ to $0.052$, representing nearly a fivefold increase. This supports the interpretation that the structural integrity of replay samples is an important factor influencing knowledge retention.

Interestingly, CardiacNet performance remains stable and slightly improves as corruption increases. This indicates a stability-plasticity trade-off: corrupted replay imposes weaker structural constraints, allowing greater adaptation to the target task, but at the cost of weaker retention of the source task. As shown in Fig.~\ref{fig:structural_corruption_analysis}, increasing structural corruption produces a monotonic increase in forgetting. This trend was consistent across all three random seeds, indicating that degradation in replay-mask structure systematically weakens source-task retention under otherwise fixed experimental conditions.

\subsection{Ablation Study}


\begin{table}[t]
\centering
\caption{Ablation study of the proposed components in the forward task order (CAMUS to CardiacNet).}
\label{tab:ablation}
\scriptsize
\setlength{\tabcolsep}{3pt}
\renewcommand{\arraystretch}{1.12}

\begin{threeparttable}
\begin{tabular*}{\columnwidth}{@{\extracolsep{\fill}}lccc@{}}
\toprule
\textbf{Method}
& \makecell{\textbf{CAMUS}\\\textbf{(After) $\uparrow$}}
& \makecell{\textbf{CardiacNet}\\\textbf{$\uparrow$}}
& \makecell{\textbf{Forgetting}\\\textbf{$\downarrow$}} \\
\midrule

Plain Replay
& $0.882 \pm 0.006$
& $0.746 \pm 0.008$
& $0.019 \pm 0.007$ \\

Weighted Sampler Only
& $0.419 \pm 0.046$
& $0.798 \pm 0.011$
& $0.483 \pm 0.046$ \\

Class Only
& $0.880 \pm 0.008$
& $0.745 \pm 0.006$
& $0.021 \pm 0.008$ \\

Boundary Only
& $0.883 \pm 0.009$
& $0.748 \pm 0.012$
& $0.018 \pm 0.009$ \\

BBR-Net (Full)
& $\mathbf{0.889 \pm 0.004}$
& $\mathbf{0.752 \pm 0.005}$
& $\mathbf{0.012 \pm 0.004}$ \\

\bottomrule
\end{tabular*}

\begin{tablenotes}
\scriptsize
\item Results are reported over three random seeds as mean $\pm$ standard deviation.
\end{tablenotes}
\end{threeparttable}
\end{table}



The ablation study in Table~\ref{tab:ablation} evaluates the contribution of each component in the proposed framework. Plain replay already provides a strong baseline, achieving a CAMUS retention score of $0.882 \pm 0.006$ and reducing forgetting to $0.019 \pm 0.007$. This indicates that replay itself is effective when the source task provides stable anatomical priors. However, the proposed Full BBR-Net further improves the overall balance between source-task retention and target-task adaptation, achieving the highest CAMUS retention ($0.889 \pm 0.004$), the best CardiacNet performance ($0.752 \pm 0.005$), and the lowest forgetting ($0.012 \pm 0.004$).

Weighted Sampler Only performs poorly in terms of retention, with CAMUS performance dropping to $0.419 \pm 0.046$ and forgetting increasing to $0.483 \pm 0.046$, despite achieving the highest CardiacNet performance ($0.798 \pm 0.011$). This confirms that weighted sampling alone promotes adaptation to the current task but does not preserve previously learned anatomical knowledge. In contrast, the Class Only and Boundary Only variants maintain strong retention, with forgetting scores of $0.021 \pm 0.008$ and $0.018 \pm 0.009$, respectively. Boundary Only slightly outperforms Class Only, suggesting that boundary information provides a more direct cue for preserving structural consistency than class-frequency balancing alone.

Overall, these results show that replay effectiveness depends not only on revisiting previous samples, but also on how structurally informative those samples are. The best performance is obtained when boundary-aware replay and class-aware prioritization are combined, supporting the design of BBR-Net. Importantly, Weighted Sampler Replay in Table~\ref{tab:forward_reverse_results} denotes weighted sampling combined with replay, whereas Weighted Sampler Only in Table~\ref{tab:ablation} isolates weighted sampling without replay memory, which explains why it adapts well to the target task but fails to preserve source-task knowledge. This distinction explains why weighted replay performs strongly in the main comparison, while weighted sampling alone fails to prevent forgetting.

\subsection{Qualitative Results}

\begin{figure*}[!t] 
\centering 
\includegraphics[width=\textwidth]{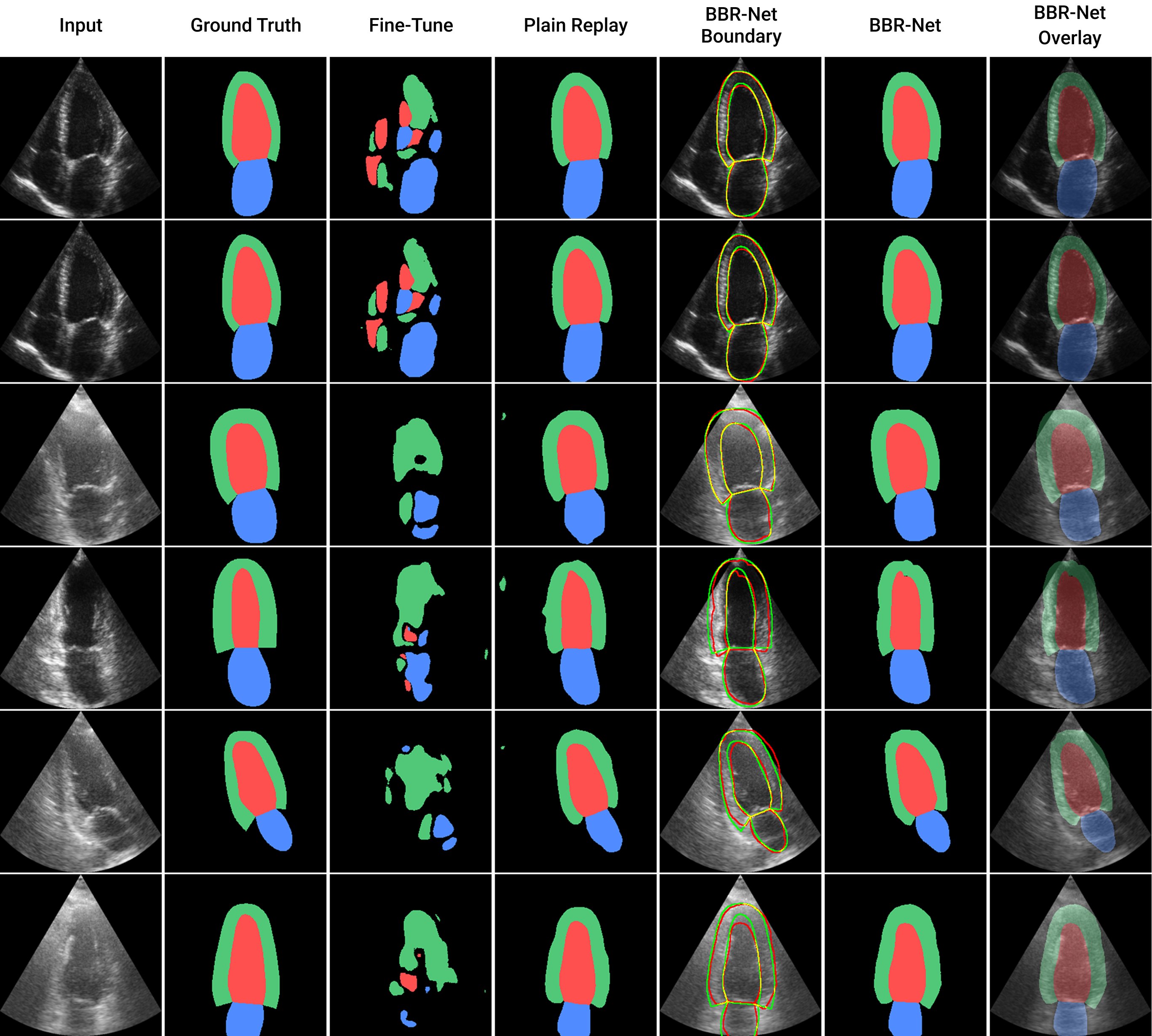} 
\caption{Qualitative results for the forward task order (CAMUS to CardiacNet). Samples are grouped by difficulty levels (best, hard, and moderate). Columns show the input image, ground truth, fine-tuning baseline, plain replay baseline, BBR-Net boundary overlay, BBR-Net prediction, and BBR-Net prediction overlay. Red, green, and blue regions denote the LV endocardium, LV epicardium, and left atrium, respectively. Yellow contours in the boundary overlay indicate predicted anatomical boundaries.}
\label{forward} 
\end{figure*}

\begin{figure*}[!t] 
\centering 
\includegraphics[width=0.72\textwidth]{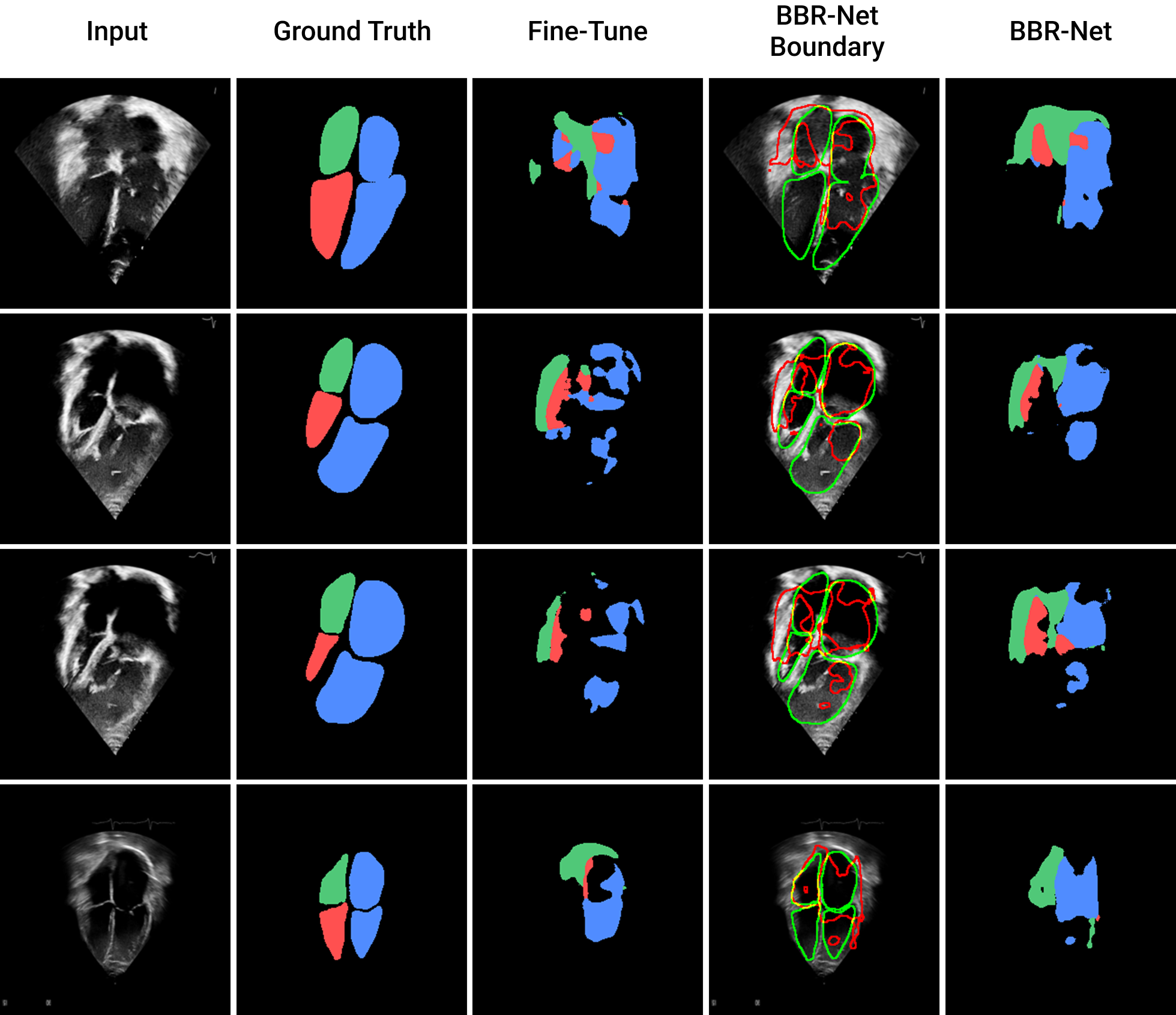} 
\caption{Comparison of best-case and hard-case samples in the forward task order (CAMUS to CardiacNet). Columns show the input image, ground truth, fine-tuning baseline, BBR-Net boundary overlay, and BBR-Net prediction. BBR-Net produces more coherent anatomical structures and smoother boundaries than the fine-tuning baseline. Red, green, and blue regions denote the LV endocardium, LV epicardium, and left atrium, respectively. Yellow contours indicate predicted anatomical boundaries.}
\label{forward_best_hard} 
\end{figure*}

\begin{figure*}[!t] 
\centering 
\includegraphics[width=0.85\textwidth]{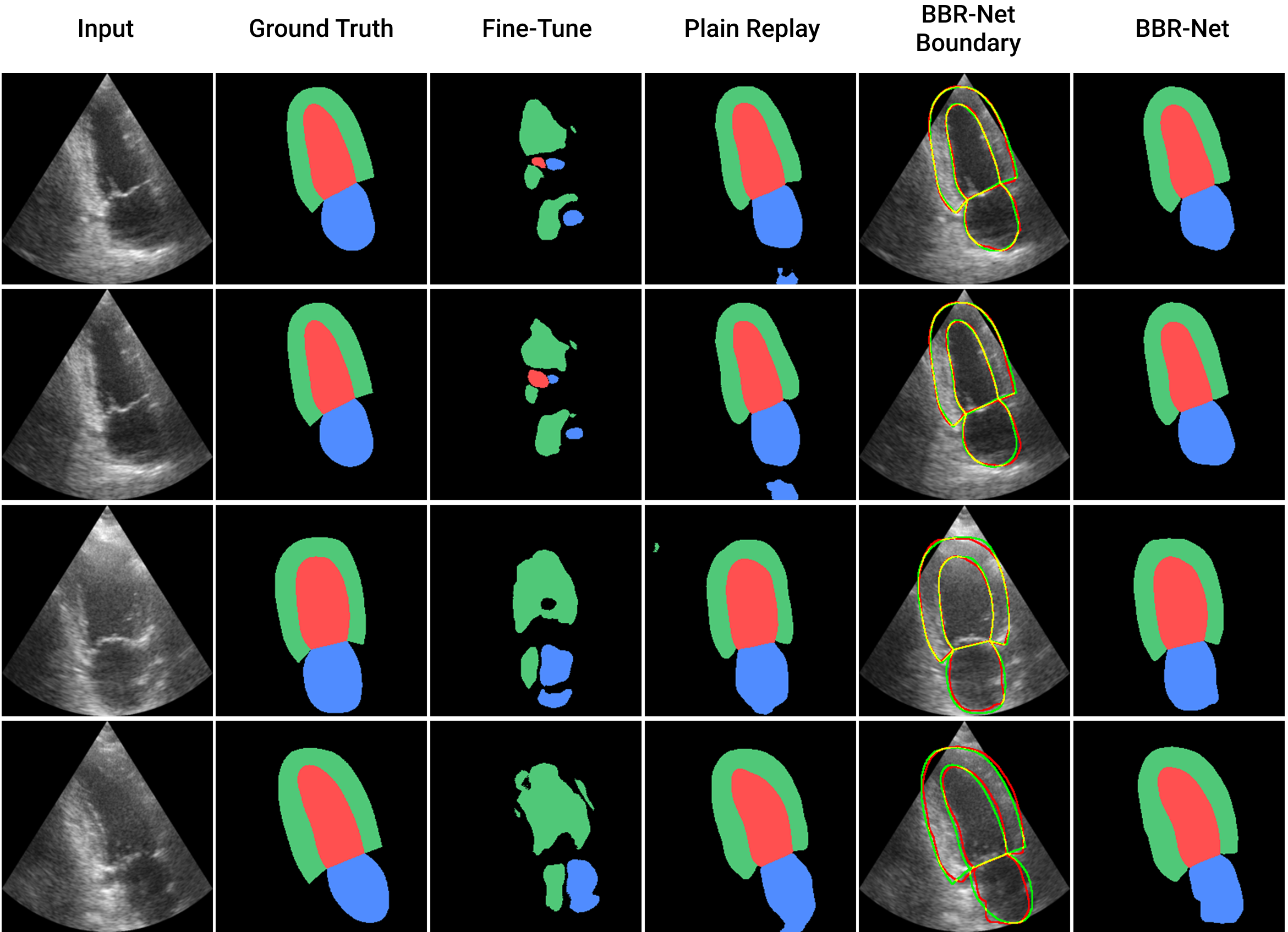} 
\caption{Qualitative results for the reverse task order (CardiacNet to CAMUS). Columns show the input image, ground truth, fine-tuning baseline, plain replay baseline, BBR-Net boundary overlay, BBR-Net prediction, and BBR-Net prediction overlay. The reverse setting illustrates how the quality of initial structural representations affects replay behavior under domain shift. Red, green, and blue regions denote the LV endocardium, LV epicardium, and left atrium, respectively. Yellow contours in the boundary overlay indicate predicted anatomical boundaries.}
\label{reverse_order} 
\end{figure*}

\begin{figure*}[!t] 
\centering 
\centerline{\includegraphics[width=0.98\textwidth]{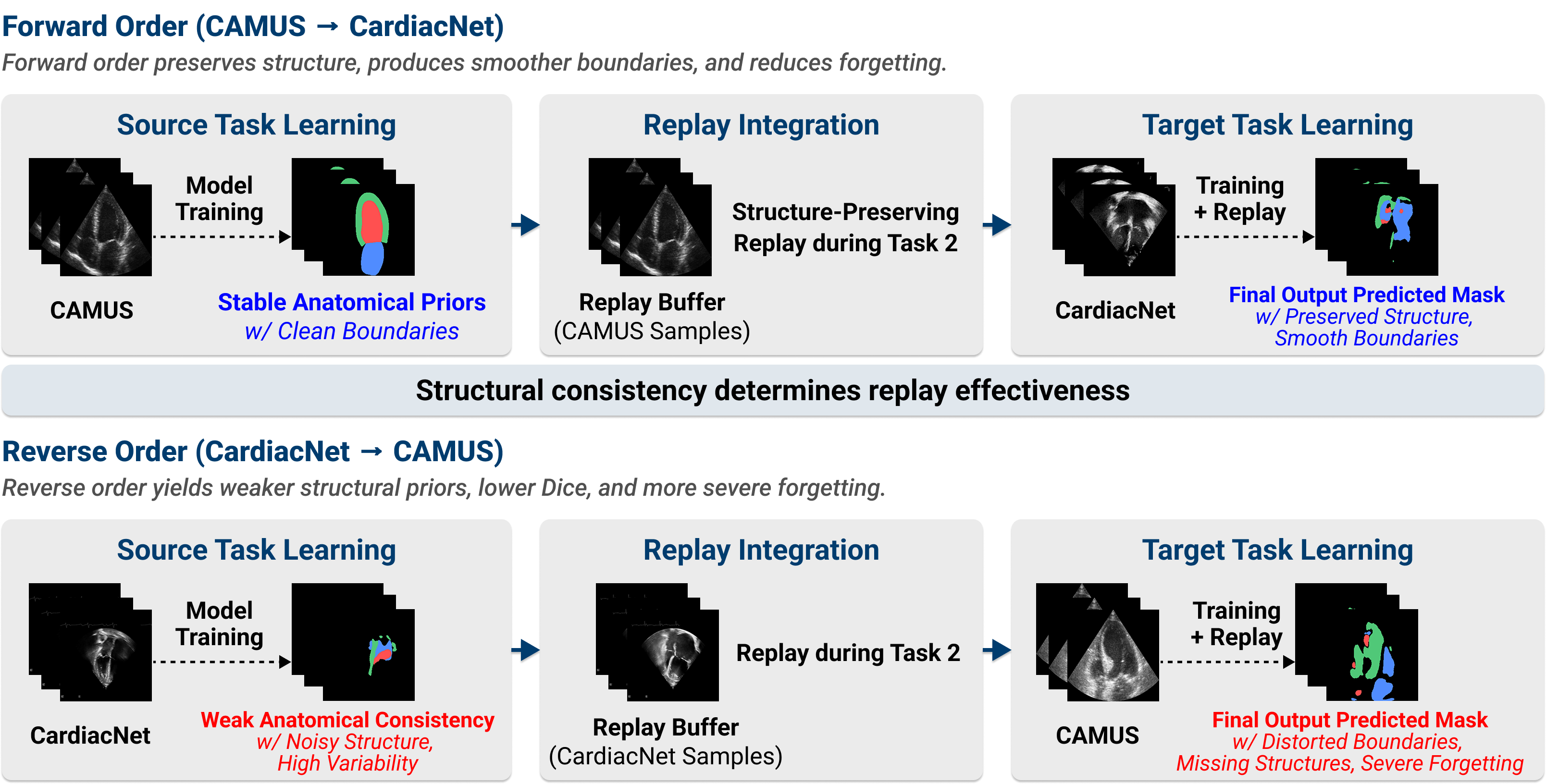}} 
\caption{Illustration of task-order effects in continual segmentation. Forward order enables stable structural learning and effective replay, while reverse order leads to weak representations and increased forgetting.}
\label{fig:task_order_effects} 
\end{figure*}

The qualitative results provide deeper insights into the behavior of the proposed BBR-Net under both forward and reverse task orders. Unlike quantitative metrics, these visualizations reveal how structural knowledge is preserved or degraded at the anatomical level.

\subsubsection{Forward Order Analysis (CAMUS to CardiacNet)}

In the forward setting (Fig.~\ref{forward} and Fig.~\ref{forward_best_hard}), the model is first trained on CAMUS, which provides clean and structurally consistent annotations. This allows the network to learn strong anatomical priors before encountering the more complex CardiacNet dataset.

\textbf{Best-case samples.}
In the first two rows, both fine-tuning and the proposed method produce visually acceptable segmentations. However, a closer inspection reveals that the fine-tuning baseline exhibits subtle boundary irregularities, including jagged edges and slight over-segmentation near class transitions. In contrast, the proposed method produces smoother and more continuous boundaries, indicating that the boundary supervision effectively regularizes the prediction space. The anatomical regions (LV endocardium, epicardium, and left atrium) are well separated with consistent spatial relationships.

\textbf{Hard samples.}
The differences become significantly more pronounced in the third and fourth rows. The fine-tuning baseline fails catastrophically, producing fragmented predictions with disconnected regions and missing anatomical structures. In several cases, entire regions (e.g., atrium or epicardium) are either partially detected or completely absent. This indicates severe forgetting of previously learned structures. In contrast, the proposed method maintains global structural coherence. Even when local errors exist, the predicted regions preserve correct topology and relative positioning. The boundaries remain aligned with anatomical contours, and the segmentation does not collapse into noisy or scattered regions. This demonstrates that replay guided by boundary and class-aware prioritization successfully reinforces structurally informative samples.

\textbf{Moderate samples.}
In the last rows, the baseline produces partially degraded predictions, with noticeable boundary shifts and inconsistent region sizes. The proposed method, however, maintains stable predictions with minor deviations, showing improved robustness under intermediate difficulty conditions. A critical observation is that the proposed method does not merely improve pixel-wise accuracy, but preserves shape consistency and anatomical plausibility. The segmentation outputs follow expected cardiac geometry, with smooth enclosing epicardial boundaries and properly nested structures. This confirms that the method learns and retains structural priors rather than relying solely on appearance cues.

\subsubsection{Forward Order: CAMUS-specific Insights}

Fig.~\ref{forward_best_hard} shows that CAMUS samples benefit strongly from replay when learned first. Since CAMUS provides clean supervision, the replay buffer stores high-quality structural exemplars. These samples act as strong anchors during subsequent training, preventing drift in decision boundaries. In hard cases, even when the baseline produces completely incoherent segmentations, the proposed method preserves the overall geometry of the heart chambers. This indicates that replay samples are not only revisited but are prioritized based on their structural importance, which is crucial for maintaining class relationships and boundary integrity. 

\subsubsection{Reverse Order Analysis (CardiacNet to CAMUS)}

The reverse setting (Fig.~\ref{reverse_order}) reveals a fundamentally different behavior and exposes the limitations of replay-based methods.

\textbf{Failure cases.}
When the model is first trained on CardiacNet, the learned representations are inherently noisy and structurally inconsistent due to variability in data quality and annotations. As a result, the replay buffer stores suboptimal or ambiguous samples. During subsequent training on CAMUS, these noisy replay samples fail to provide reliable guidance. This leads to severe catastrophic forgetting, where previously learned structures are not retained. The predictions become fragmented, with large missing regions and disconnected components. In extreme cases, the model collapses into predicting only partial structures or incorrect class assignments.

\textbf{Misleading cases.}
A particularly important observation is the presence of visually plausible but structurally incorrect predictions. In these cases, the segmentation appears reasonable at a glance, but closer inspection reveals significant boundary misalignment and incorrect topology. For example, regions may overlap incorrectly or fail to maintain proper enclosure relationships. The boundary visualizations clearly highlight these inconsistencies, showing that predicted contours deviate significantly from anatomical boundaries. This demonstrates that appearance-level similarity does not guarantee structural correctness.

Although the proposed method still outperforms the baseline in some cases, its advantage is significantly reduced compared to the forward setting. The model struggles to enforce structural consistency because the initial representations lack reliable structural priors. As a result, the replay mechanism reinforces noisy patterns rather than correcting them. These results confirm that replay effectiveness is highly dependent on the quality of the initial task. When strong structural priors are learned early (forward order), replay preserves them effectively. However, when the initial task is noisy (reverse order), replay propagates errors, leading to degraded performance. This task-order-dependent behavior is summarized in Fig.~\ref{fig:task_order_effects}, which contrasts stable forward-order replay with degraded reverse-order replay under noisy initial representations.

\subsubsection{Overall Insights}

The qualitative analysis reveals three critical insights: First, catastrophic forgetting manifests not only as performance drop but as structural collapse, where anatomical relationships are lost and predictions become fragmented. Second, the proposed method successfully mitigates this issue in the forward setting by enforcing boundary consistency and prioritizing structurally informative samples. Third, replay-based CL is closely related to the quality of stored representations. Without stable structural priors, replay cannot guarantee knowledge retention and may even amplify errors. These findings support that structure-aware learning can be important for reliable continual medical image segmentation, particularly under strong domain shift.

\section{Discussion}
\label{sec:discussion}
The results, including the structural corruption analysis, suggest that CL performance in medical image segmentation depends in part on the preservation of structural representations, rather than appearance similarity alone. This section critically examines these findings in relation to existing CL paradigms and their underlying assumptions.

\subsection{Forward Order: Structural Priors Enable Effective Replay}  
In the forward setting (CAMUS to CardiacNet), the proposed method achieves near the performance of the joint-training reference, with minimal forgetting ($\mathcal{F}=0.009$). This behavior aligns with the fundamental premise of replay-based methods, which assume that revisiting past samples approximates joint training~\cite{rebuffi2017icarl, chaudhry2019tiny}. However, our results refine this assumption by suggesting that replay is more effective when the stored samples encode stable and consistent structure.

Unlike standard replay, which treats all samples uniformly, the proposed method prioritizes boundary information and class balance. The ablation results (Table~\ref{tab:ablation}) indicate that boundary-aware replay and class balancing contribute to retention, while the forward-order results show that the full BBR-Net provides the best overall balance between source-task retention and target-task adaptation. This finding challenges the prevailing emphasis on appearance-based or uncertainty-driven sampling strategies, which assume that difficult samples are inherently more informative. Instead, our results suggest that structural informativeness is a more reliable criterion for knowledge retention.

This observation is consistent with recent analyses in medical CL~\cite{wu2024continual, qazi2024medicalclsurvey}, which highlight the limitations of appearance-based representations under domain shift. However, unlike prior works that primarily identify this issue, our results provide empirical validation by demonstrating that explicitly modeling structural cues leads to near-optimal retention. These findings extend the theoretical understanding of replay-based CL: effective memory is not defined by diversity or difficulty alone, but by its ability to preserve structurally meaningful information.

\subsection{Reverse Order: Breakdown of Replay under Structural Inconsistency}  
In the reverse setting (CardiacNet to CAMUS), the behavior changes fundamentally. Plain replay and weighted sampler replay retain CardiacNet knowledge relatively well, whereas boundary-aware replay and Full BBR-Net collapse to near fine-tuning levels. This contrast reveals that structure-aware replay is not universally beneficial; it depends on whether the initial structural representations are reliable.

Existing methods such as EWC~\cite{kirkpatrick2017ewc}, iCaRL~\cite{rebuffi2017icarl}, and dynamic memory approaches~\cite{perkonigg2021dynamic} assume that stored knowledge remains compatible with future tasks. Our results reveal a critical limitation of this assumption: when initial representations are learned from noisy or structurally inconsistent data, replay propagates errors rather than correcting them. This behavior is not adequately explained by optimization or memory size, but by the incompatibility between the stored representations and the evolving feature space. Even advanced replay variants (Table~\ref{tab:forward_reverse_results}) fail to recover performance, indicating that the issue is structural rather than algorithmic. These findings introduce an important nuance to CL theory: replay effectiveness is conditional on representation quality. Without stable structural priors, memory-based methods cannot guarantee knowledge retention.

\subsection{Controlled Structural Corruption Confirms the Role of Structure}

The controlled structural corruption experiment provides additional empirical support for the central hypothesis of this study. Unlike the forward and reverse task-order comparison, this experiment keeps the dataset order, model architecture, training schedule, and memory size fixed, while only degrading the structural reliability of source-task replay masks. This allows the effect of structural consistency to be examined more directly. The results show a clear monotonic trend: as corruption severity increases, retention on the source task decreases and forgetting increases. Specifically, CAMUS after-task performance drops from $0.890$ under clean replay to $0.848$ under severe corruption, while forgetting increases from $0.011$ to $0.052$. This pattern suggests that replay effectiveness is not determined only by the presence of memory, but is also strongly affected by the structural quality of the samples stored in that memory.

Interestingly, target-task performance on CardiacNet remains stable and slightly improves as corruption increases. This suggests that corrupted replay imposes weaker constraints on adaptation, allowing the model to fit the new task more freely. However, this comes at the cost of reduced retention. This finding provides empirical support for a structure-related stability-plasticity trade-off: clean structural replay supports stability, whereas corrupted replay weakens retention while increasing plasticity. These results are important because they address a potential alternative explanation for the forward-reverse asymmetry. One could argue that the reverse-order failure is caused mainly by dataset difficulty, noise level, or annotation quality differences between CAMUS and CardiacNet. The controlled corruption experiment reduces this ambiguity by isolating structural reliability as the manipulated factor. Therefore, the findings provide stronger empirical evidence that structural consistency is an important factor in replay effectiveness in continual medical image segmentation.

From a practical perspective, the results suggest that Weighted Sampler Replay may provide a more robust default strategy when task ordering and structural reliability are unknown, whereas structure-aware replay becomes advantageous when stable anatomical priors can be learned before substantial domain shifts occur.

\subsection{Stability-Plasticity Trade-off Revisited}  
The slight reduction in CardiacNet performance in the forward setting ($0.777$ vs $0.795$ for fine-tuning) reflects a controlled stability-plasticity trade-off. Traditional CL literature often frames this trade-off as a balance between retaining old knowledge and adapting to new tasks~\cite{parisi2019continual, delange2021continual}. Our results suggest a more nuanced interpretation. Rather than a generic trade-off, the observed behavior reflects a prioritization of structural consistency over short-term adaptation. The proposed method sacrifices a small degree of plasticity to preserve globally coherent anatomical representations, which ultimately leads to better long-term performance. This redefines the stability-plasticity trade-off in segmentation: preserving structure may be more important than maximizing immediate task performance.

\subsection{Comparison with Existing Approaches}  
Compared to prior work, the proposed method introduces a distinct perspective. Regularization-based methods such as EWC \cite{kirkpatrick2017ewc} focus on parameter importance, but do not account for spatial structure. Replay-based methods such as Incremental Classifier and Representation Learning (iCaRL) \cite{rebuffi2017icarl} emphasize sample diversity, while recent segmentation approaches (e.g., RECALL \cite{maracani2021recall} and RCIL \cite{zhang2022rcil}) rely on feature alignment or generative replay. 

Our findings both align with and challenge these approaches. On one hand, they confirm that replay is a powerful mechanism for mitigating forgetting. On the other hand, they reveal that existing methods overlook a critical factor: structural consistency. This gap explains why many approaches perform well in controlled settings but degrade under strong domain shifts.

The key contribution of this work lies in demonstrating that structure-aware replay not only improves performance but also exposes a fundamental limitation of current CL methods under domain shift. From a theoretical perspective, this study shifts the focus of CL from distribution preservation to structural preservation. It suggests that future methods should explicitly model geometric and anatomical constraints rather than relying solely on feature-level alignment. From a practical standpoint, the findings have direct implications for clinical deployment. In real-world settings, models are often updated sequentially with data from different institutions or devices. Ensuring structural consistency is therefore critical for maintaining reliable segmentation performance. From an application perspective, the results indicate that naive replay strategies may fail in heterogeneous clinical environments. Structure-aware mechanisms provide a more robust alternative for long-term deployment.


Despite the promising results, several limitations must be acknowledged. First, the method exhibits strong sensitivity to task order. While near-optimal retention is achieved in the forward setting, performance degrades significantly in the reverse setting. This indicates that the approach relies on the availability of stable structural priors during early training. Second, the method assumes that boundary information reliably encodes anatomical structure. While this holds for cardiac segmentation, it may not generalize to tasks where boundaries are ambiguous or poorly defined. Third, the evaluation is limited to a two-task setting. Extending the framework to longer task sequences may introduce cumulative error propagation, increased memory requirements, and more complex interactions between tasks. Finally, the study focuses on a single anatomical domain. The generalizability of structure-aware replay to other modalities or organs remains to be validated.

Future research should explore hybrid approaches that combine structure-aware replay with representation alignment or generative modeling. Developing methods that can learn robust structural priors from noisy data is particularly important for addressing reverse-order failure. Additionally, extending the framework to multi-task and multi-modal settings will be critical for real-world deployment. Finally, investigating task-order invariant learning strategies may provide a more principled solution to the limitations identified in this study.

\section{Conclusion and Future Work}
\label{conclusion}

This study investigated a central question in continual medical image segmentation: when and why replay-based methods succeed or fail under domain shift. Specifically, the study aimed to (i) evaluate the effectiveness of replay for mitigating catastrophic forgetting, (ii) examine the role of structural consistency in preserving anatomical knowledge, (iii) analyze the influence of task order on CL dynamics, and (iv) examine whether replay effectiveness is affected by the structural reliability of stored samples. The experimental results suggest that CL performance is strongly influenced by the quality of structural representations, rather than appearance similarity alone. In the forward task order (CAMUS to CardiacNet), the proposed BBR-Net retains source-task performance close to the offline joint-training reference, reducing forgetting to a negligible level while maintaining competitive target-task performance. The ablation study further indicates that boundary-aware prioritization contributes to retention, particularly when combined with class-aware replay selection, indicating that structurally informative samples are substantially more effective than appearance-based or frequency-based replay strategies. These findings show that preserving anatomical boundaries enables the replay buffer to retain clinically meaningful representations over sequential learning stages. 

More importantly, the reverse task order (CardiacNet to CAMUS) exposes a fundamental limitation of replay-based CL. When the initial task contains noisy and structurally inconsistent annotations, the learned representations fail to establish stable anatomical priors. Under this condition, replay becomes ineffective and may even propagate structurally unreliable information, resulting in severe forgetting despite the presence of memory. This observation indicates that replay effectiveness is not universal, but strongly dependent on the compatibility between stored samples and the evolving representation space. 

To further validate this hypothesis, we conducted a controlled structural perturbation analysis in which boundary corruption was progressively introduced while keeping the dataset, architecture, training schedule, and replay configuration fixed. The results revealed a consistent degradation in source-task retention and a corresponding increase in forgetting as corruption severity increased. This experiment provides controlled empirical evidence that replay performance is strongly affected by the structural reliability of stored representations. Importantly, the findings demonstrate that the degradation observed in the reverse task order is unlikely to be explained solely by dataset difficulty or appearance variation, and appears closely linked to the loss of stable structural information during replay construction.

The broader significance of this work lies in reframing continual medical image segmentation as a problem of structural preservation rather than appearance preservation. Existing CL approaches largely focus on maintaining feature distributions or parameter stability, whereas our results show that anatomical consistency, particularly boundary integrity, plays a more decisive role in long-term knowledge retention. From a theoretical perspective, these findings suggest that future CL frameworks should move beyond distribution alignment and incorporate explicit structural constraints. From a practical perspective, the study highlights the importance of preserving clinically meaningful anatomical information for reliable deployment in real-world multi-center and multi-domain clinical environments.

Despite these contributions, several limitations remain. The current study focuses on sequential learning across two datasets and relies primarily on replay-based mechanisms. Although the controlled perturbation experiment strengthens the structural consistency hypothesis, the analysis remains limited to cardiac ultrasound segmentation and relatively short task sequences. In addition, the proposed framework does not explicitly enforce representation-level anatomical invariance, which may be necessary under severe structural noise and heterogeneous domain shifts.

Future research should therefore focus on developing structure-constrained representation learning strategies capable of preserving anatomical priors even under noisy supervision. Integrating structure-aware replay with generative modeling, uncertainty estimation, or domain adaptation techniques may further improve robustness across heterogeneous clinical settings. Extending the framework to longer task sequences, multi-modal imaging scenarios, and large-scale real-world clinical datasets will also be essential for improving generalization and practical applicability.

In summary, this work advances the understanding of CL by demonstrating that structural consistency, rather than appearance similarity, is the key factor influencing replay effectiveness in medical image segmentation. By introducing a structure-aware perspective and validating it through controlled structural perturbation analysis, this study not only improves continual segmentation performance in favorable settings but also exposes fundamental limitations of existing replay-based methods, providing a foundation for the next generation of structurally informed CL approaches in medical imaging.

\section*{CRediT authorship contribution statement}
\textbf{Zahid Ullah:} Conceptualization, Methodology, Software, Formal analysis, Investigation, Data curation, Writing - original draft, Writing - review \& editing. \textbf{Sieun Choi:} Conceptualization, Methodology, Formal analysis, Writing - original draft, Writing - review \& editing. \textbf{Jihie Kim:} Formal analysis, Investigation, Supervision, Project administration, Project management.

\section*{Declaration of Competing Interests}
The authors declare that they have no known competing financial interests or personal relationships that could have appeared to influence the work reported in this paper.


\section*{Data avalailability}
The used datasets are publicly available.

\section*{Acknowledgment}
{This work was partly supported by the Institute of Information \& Communications Technology Planning \& Evaluation(IITP)-ITRC(Information Technology Research Center) grant funded by the Korea government(MSIT)(IITP-2026-RS-2020-II201789),
 and the Artificial Intelligence Convergence Innovation Human Resources Development(IITP-2026-RS-2023-00254592) supervised by the IITP(Institute for Information \& Communications Technology Planning \& Evaluation).}

\bibliographystyle{cas-model2-names}
\bibliography{cas-refs.bib}

\end{document}